%% file: main.tex
\documentclass{article}

\usepackage[preprint]{neurips_2025}

\usepackage[utf8]{inputenc} 
\usepackage[T1]{fontenc}    
\usepackage[hidelinks]{hyperref}\usepackage{url}            
\usepackage{booktabs}       
\usepackage{amsfonts}       
\usepackage{nicefrac}       
\usepackage{microtype}      
\usepackage{xcolor}         
\usepackage{xspace}
\usepackage{amsmath}
\usepackage{amssymb}
\usepackage{mathtools}
\usepackage{amsthm}
\usepackage{hyperref}
\usepackage{microtype}
\usepackage{graphicx}
\usepackage{booktabs}
\usepackage{formatting/macros}
\usepackage{formatting/results}
\usepackage{float}
\usepackage{subfig}
\usepackage[small]{complexity}
\usepackage{etoolbox,xspace}
\usepackage{soul}
\usepackage{wrapfig}
\usepackage{pifont}
\usepackage{multirow}
\usepackage{enumitem}
\usepackage{algorithm}
\usepackage{algorithmic}
\usepackage{graphicx}

\robustify\ComplexityFont
\apptocmd\ComplexityFont{\xspace}{}{}

\title{Foresight: Adaptive Layer Reuse for Accelerated and High-Quality Text-to-Video Generation}

\author{%
  Muhammad Adnan\thanks{A large part of this work was done while the author was interning with d-Matrix.} \\
  The University of British Columbia\\
  \texttt{adnan@ece.ubc.ca} \\
  \And
  Nithesh Kurella\\
  d-Matrix\\
  \texttt{nitheshk@d-matrix.ai} \\
  \AND
  Akhil Arunkumar \\
  d-Matrix \\
  \texttt{aarunkumar@d-matrix.ai} \\
  \And
  Prashant J. Nair \\
  The University of British Columbia \\
  \texttt{prashantnair@ece.ubc.ca} \\
}

\begin{document}

\maketitle

\begin{abstract}
  \input{./body/abstract}
\end{abstract}

\input{./body/introduction}
\input{./body/related_work}
\input{./body/methodology}
\input{./body/evaluation}
\input{./body/limitations}
\input{./body/conclusion}

\bibliography{references}
\bibliographystyle{plainnat}

\newpage

\appendix

\input{./body/appendix}


\end{document}

%% file: body/abstract.tex
Diffusion Transformers (DiTs) achieve state-of-the-art results in text-to-image, text-to-video generation, and editing. However, their large model size and the quadratic cost of spatial-temporal attention over multiple denoising steps make video generation computationally expensive. Static caching mitigates this by reusing features across fixed steps but fails to adapt to generation dynamics, leading to suboptimal trade-offs between speed and quality.

We propose Foresight, an adaptive layer-reuse technique that reduces computational redundancy across denoising steps while preserving baseline performance. Foresight dynamically identifies and reuses DiT block outputs for all layers across steps, adapting to generation parameters such as resolution and denoising schedules to optimize efficiency. Applied to OpenSora, Latte, and CogVideoX, Foresight achieves up to \latencyimprv end-to-end speedup, while maintaining video quality.
The source code of Foresight is available at \href{https://github.com/STAR-Laboratory/foresight}{https://github.com/STAR-Laboratory/foresight}.

%% file: body/introduction.tex
\section{Introduction}
\label{sec:introduction}

Diffusion Transformers (DiTs)~\citep{dit1,dit2} are the state-of-the-art architecture for text-to-image~\citep{pixart} and text-to-video~\citep{sora,mochi,latte,textanimator} generation. By leveraging the scaling properties of self-attention~\citep{attention}, DiTs enable high-fidelity video synthesis. However, this comes at a high computational cost. Self-attention has quadratic complexity $\mathcal{O}(L^2)$ in the token length $L$, which increases with both spatial resolution (more patches per frame) and temporal extent (more frames). Combined with the \emph{tens} of denoising steps in typical diffusion pipelines, this leads to prohibitive inference latency for high-resolution or long-duration videos.

Inference latency in diffusion models is further increased by classifier-free guidance (CFG)~\citep{cfg} and full-precision weights. Techniques like pruning~\citep{pruning}, quantization~\citep{quantization}, distillation~\citep{distillation}, and architectural optimizations~\citep{snapfusion,mobilediffusion,viditq,structuralpruning,tokenmerging} can reduce model size and compute cost. However, they often require retraining or hardware-specific tuning, making them impractical for deploying large, pretrained DiTs across diverse hardware environments.

\begin{figure*}[t]
  \centering
  \subfloat[Qualitative]{
  \begin{minipage}[t]{0.61\textwidth}
  \centering
  \includegraphics[width=\textwidth]{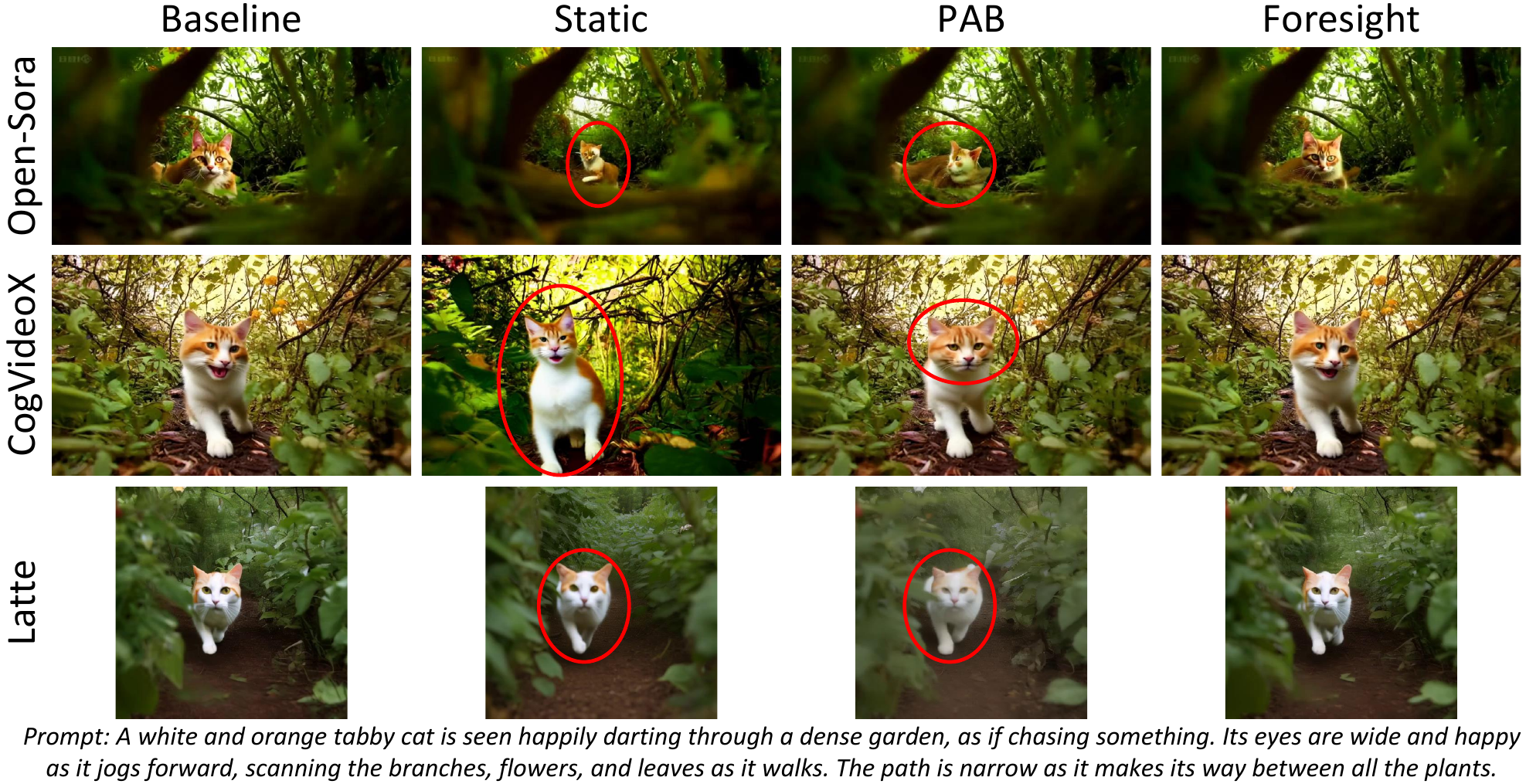}
  \end{minipage}
  }
  \hfill
  \subfloat[Quantitative]{
  \begin{minipage}[t]{0.36\textwidth}
      \centering
      \includegraphics[width=\textwidth]{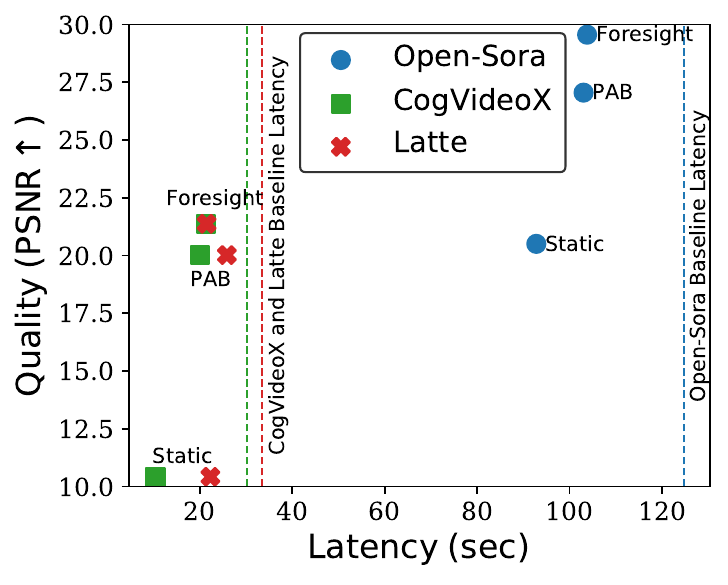}
  \end{minipage}
  }
  \caption{We compare the effectiveness of Adaptive Layer Reuse (\foresight) with prior state-of-the-art static techniques. The \textcolor{red}{red circle} indicates the frame variations in prior techniques. Videos are generated using OpenSora~\cite{opensora} at 720p, Latte~\cite{latte} at 512×512, and CogVideoX~\cite{cogvideox} at 480×720, all with a fixed length of 2 seconds. We report average inference time on a single A100 GPU using Open-Sora prompts.
}
\label{fig:motivation}
\end{figure*}

Feature caching offers a complementary strategy for accelerating diffusion by reusing intermediate activations across adjacent denoising steps. Early methods for U-Net models cache skip-connection outputs between steps $t$ and $t{-}1$~\citep{deepcache,blockcache}, while recent work extends this idea to DiTs~\citep{deltadit,tgate,fora,ditfastattn,tokencache,adacache,teacache}. However, these approaches apply a \emph{static}, uniform reuse policy across all layers and steps, which yields limited speedup and often degrades video quality and temporal coherence. As shown in Figure~\ref{fig:motivation}, static methods~\citep{deltadit, tgate, pab} introduce visible artifacts (highlighted in red), driven by a latency-first design that ignores layer- and step-specific reuse dynamics.

To better balance inference speed and generation quality, we analyze reuse patterns in DiT blocks and find that static reuse fails to preserve critical updates. Our analysis reveals two key findings: (1) the potential for reuse varies significantly across both layers and denoising steps, and (2) even minor changes in generation conditions, such as prompt content, resolution, or step count, can dramatically alter activation behavior. Figure~\ref{fig:reuse}(Left) quantifies the layer-wise Mean Squared Error (MSE) of spatial DiT features between successive timesteps in OpenSora, revealing pronounced layer-wise heterogeneity. Figure~\ref{fig:reuse}(Middle, Right) further highlights resolution-dependent reuse variation within the same layer. Motivated by these observations, we propose \foresight, a \emph{training-free}, per-layer adaptive reuse scheduler. \foresight~ \emph{dynamically} decides, at runtime, whether to recompute or reuse each block's activation, thereby providing accelerated inference while maintaining video quality.

Our comprehensive experiments on NVIDIA A100 GPUs for text-to-video generation benchmarks demonstrate the effectiveness of \foresight~ compared to static reuse baselines on OpenSora, Latte, and CogVideoX. \foresight~ achieves up to \latencyimprv end-to-end inference speedup, with no additional training, while preserving video quality.

%% file: body/related_work.tex
\section{Related Work}
\label{sec:related_work}

\subsection{Diffusion Transformers}
Diffusion Models (DMs)~\citep{diffusion1,pixart} generate images and videos by iteratively denoising random noise across multiple timesteps. Early models used convolutional U-Net backbones~\citep{diff_unet1,diff_unet2}, achieving strong results but with limited scalability. Diffusion Transformers (DiTs)~\citep{dit1,dit2} replace convolutions with self-attention, offering better scalability and generation quality. DiTs now set state-of-the-art benchmarks in text-to-image~\citep{textanimator} and text-to-video generation~\citep{video_gen}. Notably, Sora~\citep{sora} demonstrated high-quality video synthesis, motivating open-source projects like OpenSora~\citep{opensora}, Latte~\citep{latte}, CogVideoX~\citep{cogvideox}, and Mochi~\citep{mochi}. However, iterative denoising and spatial-temporal attention introduce high inference latency, limiting practical deployment.

\begin{figure*}[t]
  \centering
  \includegraphics[width=1\columnwidth]{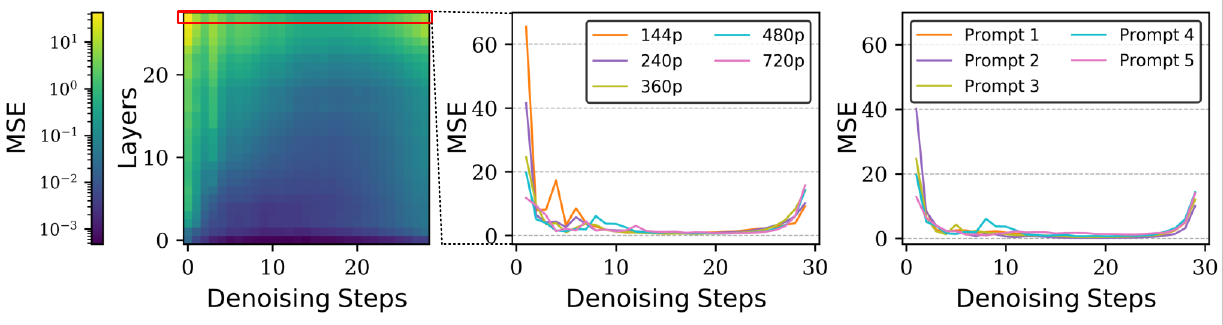}
  \caption{Mean squared error (MSE) between consecutive Spatial DiT block outputs in OpenSora~\citep{opensora} during the generation of a 2-second, 480p video. Lower MSE values indicate greater similarity and thus higher reuse potential. \textbf{(Left)} Heatmap across layers and denoising steps reveals distinct, layer-specific reuse patterns. \textbf{(Middle)} Variation of MSE for \emph{Layer 28} across different resolutions. \textbf{(Right)} Variation of MSE for Layer 28 across different text prompts. Note that Temporal DiT blocks demonstrate similar reuse trends as Spatial DiT blocks.
}
\label{fig:reuse}
\end{figure*}

\subsection{Accelerating Diffusion Transformers}
Efforts to accelerate diffusion models fall into two main categories: (1) \emph{reducing denoising steps} and (2) \emph{speeding up the denoising network}. The first reduces inference latency by shortening the sampling schedule. Deterministic methods like DDIM~\citep{ddim} preserve quality with fewer steps, while higher-order solvers such as DPM-Solvers~\citep{dpm,dpm+} improve efficiency through advanced numerical integration. Rectified Flow~\citep{rlflow} enhances ODE-based sampling for faster inference. Step distillation~\citep{guided_distillation,progressive_distillation} compresses multi-step models into fewer-step variants. Consistency Models~\citep{consistencymodels} enable single-step generation by mapping noise directly to data while maintaining consistency across timesteps.

The second direction accelerates the denoising network itself using techniques such as quantization~\citep{q-diffusion,viditq,post-quantization}, token reduction~\citep{tokenmerging,tokensimilarity}, knowledge distillation~\citep{snapfusion}, and pruning~\citep{patch-diffusion}. More recently, training-free caching methods have gained attention for reusing intermediate activations across timesteps. Methods like DeepCache~\citep{deepcache}, BlockCache~\citep{blockcache}, and Faster Diffusion~\citep{fasterdiffusion} achieve speedups via feature reuse but are designed for U-Net architectures, limiting their applicability to transformer-based models.

To overcome the limitations of U-Net, recent work adapts caching techniques to Diffusion Transformers (DiTs). Methods like $\Delta$-DiT~\citep{deltadit}, T-GATE~\citep{tgate}, and FORA~\citep{fora} apply caching to image synthesis. For video, PAB~\citep{pab} hierarchically broadcasts attention, DiTFastAttn~\citep{ditfastattn} reduces redundancy across condition and timestep dimensions, and ToCA~\citep{toca} selectively caches tokens. TeaCache~\citep{teacache} adds timestep embedding awareness to caching. Other methods~\citep{adacache,adaspa,fastercache,flexidit} explore different reuse granularities. However, they all apply static reuse uniformly across layers and steps. Notably, these techniques prioritize speedup often at the cost of degraded output quality.

%% file: body/methodology.tex
\section{Foresight: From Static to Adaptive Layer Reuse}
\label{sec:methodology}

We begin by outlining background on diffusion models and static caching for inference acceleration. We then analyze the limitations of static reuse in Diffusion Transformers (DiTs), motivating \foresight{}, a training-free framework for adaptive layer reuse.

\subsection{Preliminary}
\label{subsec:diffusion_models}

Diffusion models generate images and videos through two complementary processes. A forward process that gradually introduces noise into clean data, and a reverse process that iteratively denoises to reconstruct data conditioned on input prompts. Starting from an initial clean sample $\mathbf{x}_{0}$, the forward process adds Gaussian noise over $T$ timesteps using a variance schedule $\{\beta_t\}_{t=1}^{T}$:

\begin{equation}
\mathbf{x}_{t}= \sqrt{1-\beta_t}\,\mathbf{x}_{t-1} + \sqrt{\beta_t}\,\mathbf{z}_t,
\qquad
\mathbf{z}_t \sim \mathcal{N}(\mathbf{0},\mathbf{I}), 
\quad t = 1,\dots,T
\label{eq:forward}
\end{equation}

As $t\!\to\!T$, the distribution of $\mathbf{x}_{t}$ approaches $\mathcal{N}(\mathbf{0},\mathbf{I})$. The reverse denoising process seeks to recover the original sample $\mathbf{x}_0$ by iteratively estimating:

\begin{equation}
p_{\theta}(\mathbf{x}_{t-1}\mid \mathbf{x}_{t})
= \mathcal{N}\!\bigl(\mathbf{x}_{t-1};\,\mu_{\theta}(\mathbf{x}_{t},t),\,\Sigma_{\theta}(\mathbf{x}_{t},t)\bigr)
\label{eq:reverse}
\end{equation}

where $\mu_{\theta}$ and $\Sigma_{\theta}$ are the mean and variance predicted by a denoising neural network ($p$) that is parameterized by $\theta$.

State-of-the-art text-to-video models adopt Spatial–Temporal DiT (ST-DiT) layers that alternate spatial and temporal transformer blocks:
\[
\mathrm{DiT_{spatial}}=\{f_{\mathrm{SA}},\,f_{\mathrm{CA}},\,f_{\mathrm{MLP}}\},\qquad
\mathrm{DiT_{temporal}}=\{f_{\mathrm{TA}},\,f_{\mathrm{CA}},\,f_{\mathrm{MLP}}\}
\]
where $f_{\mathrm{SA}}$ and $f_{\mathrm{TA}}$ apply spatial and temporal self-attention, and $f_{\mathrm{CA}}$ performs cross-attention with text tokens. At timestep $t$, the input is a token sequence obtained by flattening all $H\times W$ latent patches in each of the $F$ frames:
\[
\mathbf{x}_{t}= \bigl\{x^{(f)}_{1:H\times W}\bigr\}_{f=1}^{F}
\]

\subsection{Static Layer Reuse in Diffusion Models}

Recent methods accelerate DiTs via \emph{static caching}~\citep{pab,deltadit,tgate,teacache,tokencache}.  
For a model with $L$ layers and $T$ denoising steps, let $\mathbf{x}_{t}^{l}$ denote the output of layer $l$ at step $t$, comprising spatial and temporal components  
$\bigl\{\mathbf{x}_{\mathrm{spatial}}^{\,l}(t),\,\mathbf{x}_{\mathrm{temporal}}^{\,l}(t)\bigr\}$. 
These activations are stored in a cache $\mathcal{C}$:

\begin{equation}
\mathcal{C}\!\bigl(\mathbf{x}_{t}^{l}\bigr)
\;\leftarrow\;
\bigl\{\mathbf{x}_{\mathrm{spatial}}^{\,l}(t),\,\mathbf{x}_{\mathrm{temporal}}^{\,l}(t)\bigr\},
\qquad
l=1,\dots,L,\; t=1,\dots,T
\label{eq:cache_write}
\end{equation}

All layers are computed and cached at a chosen step $t_{1}$. For the next $N$ timesteps, the model reuses those activations:

\begin{equation}
\mathbf{x}_{t}^{l}= \mathcal{C}\!\bigl(\mathbf{x}_{t_{1}}^{l}\bigr),
\qquad
l=1,\dots,L,\;
t=t_{1}+1,\dots,t_{1}+N
\label{eq:cache_read}
\end{equation}

This reuse significantly reduces computational costs by skipping redundant calculations across spatial and temporal DiT blocks for $N$ consecutive steps. Once the reuse window expires (at timestep $t_1 + N + 1$), all layers are recomputed, and the cache is refreshed with the latest outputs $\mathcal{C}(\mathbf{x}_{t_1 + N + 1}^l)$, enabling reuse in subsequent steps.

After step $t_{1}+N$, every layer is recomputed, the cache refreshed, and the cycle can repeat. Caching strategies can be applied at different granularities: fine-grained reuse (e.g., individual attention or MLP blocks) or coarse-grained reuse (entire DiT blocks). Prior methods, such as~\citep{pab}, emphasize fine-grained reuse. In contrast, our approach investigates coarse-grained reuse at the DiT block level, motivated by the observation that adjacent-step features exhibit significant similarity, as illustrated in Figure~\ref{fig:reuse}.

\begin{figure*}[t]
  \centering
  \subfloat[Prompt Dynamics]{
  \begin{minipage}[t]{0.64\textwidth}
      \centering
      \includegraphics[width=\textwidth]{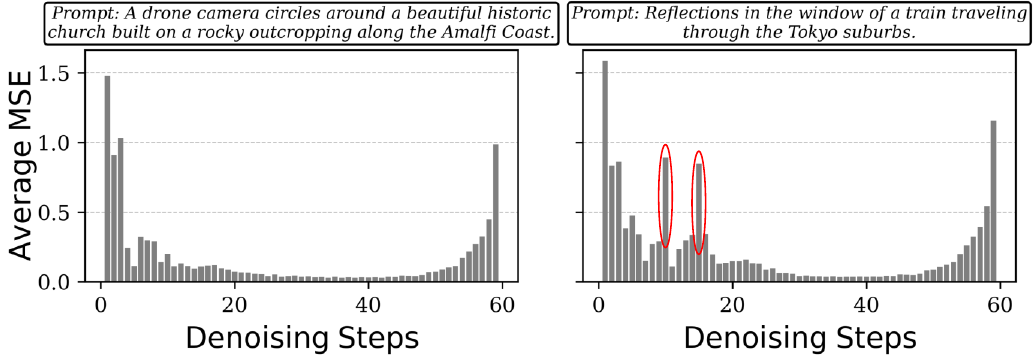}
      \vspace{-4ex}
      \label{fig:prompt_dynamics}
  \end{minipage}
  }
  \hfill
  \subfloat[Layer Dynamics]{
  \begin{minipage}[t]{0.33\textwidth}
  \centering
  \includegraphics[width=\textwidth]{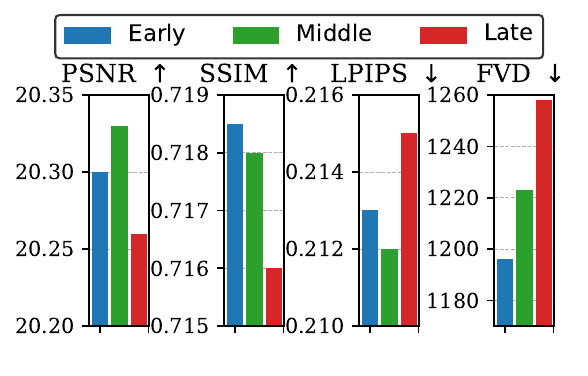}
  \vspace{-3ex}
  \label{fig:layer_dynamics}
  \end{minipage}
  }
  \caption{\textbf{(a)} Prompt-dependent dynamics of intermediate spatial features (Spatial DiT block outputs) across layers. Prompts featuring rapid scene changes exhibit sharper variations between consecutive steps, indicating higher sensitivity and less reuse potential. The \textcolor{red}{red circles} indicate these variations. \textbf{(b)} Sensitivity analysis across DiT layers grouped into early, middle, and late stages. Applying static reuse ($N=1$) within each group shows that later-stage layers disproportionately contribute to quality degradation under static caching. All experiments use the OpenSora model for 2-second, 720p videos with prompts selected from the OpenSora prompt set.}
\end{figure*}

\subsection{Limitations of Static Reuse}
\label{subsec:limitations}

\begin{figure*}[t]
  \centering
  \includegraphics[width=1\columnwidth]{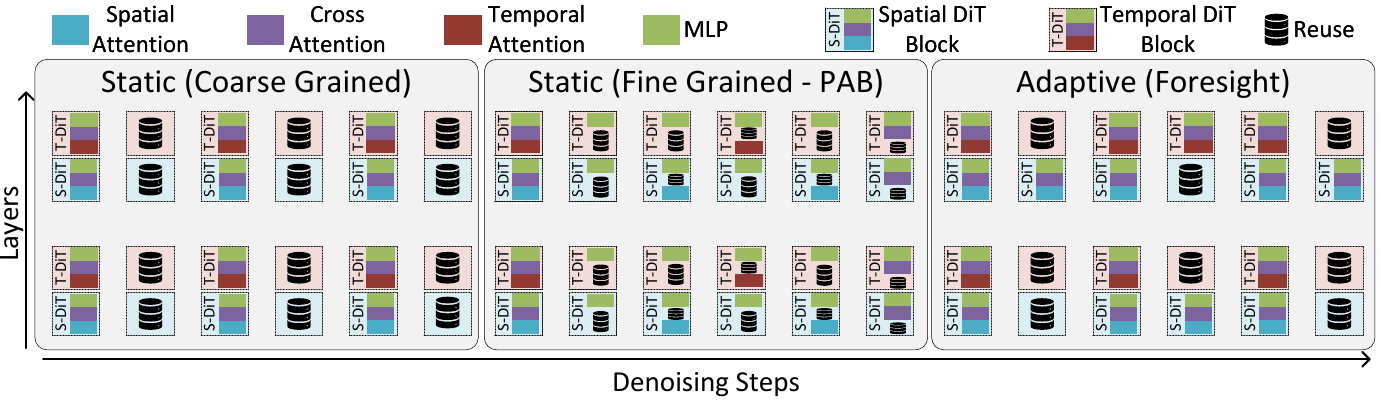}
   \caption{Overview of \foresight{} as compared to Static Reuse Techniques. Both Coarse-Grained and Fine-Grained~\citep{pab} techniques have a preset reuse strategy and do not adapt reuse dynamically with layers and denoising steps.}
   \label{fig:foresight_v1}
\end{figure*}

Reuse suitability in text-to-video diffusion varies along three key axes: prompt, layer, and configuration dynamics. Prompts differ in visual complexity. For instance, some prompts induce static scenes while others cause rapid changes. This leads to significant variation in feature similarity across timesteps (Figure~\ref{fig:prompt_dynamics}). Static reuse applies uniform caching and cannot adapt to such prompt-specific behavior. Layer-wise sensitivity analysis shows that reusing late layers degrades quality most, as these layers exhibit greater feature variation (Figure~\ref{fig:layer_dynamics}, Figure~\ref{fig:reuse} (left)), yet static methods fail to account for this variation. Moreover, video configuration parameters such as resolution, length, and denoising schedule can drastically alter reuse patterns, even under the same prompt (Figure~\ref{fig:reuse} (middle); Appendix~\ref{subsec:videoconfigs}). These findings underscore the limitations of static reuse under dynamic generation conditions.

\subsection{Foresight: Adaptive Layer Reuse Framework}

To address the limitations of static reuse, we propose \foresight{}, an adaptive layer reuse framework that balances inference speed and generation quality. Figure~\ref{fig:foresight_v1} shows the overall flow, contrasting \foresight{} with coarse-grained static caching and fine-grained methods like PAB~\citep{pab}. Unlike static approaches, \foresight{} makes dynamic reuse decisions through a two-phase process: a \emph{warmup phase} and a \emph{reuse phase}, driven by layer-specific reuse thresholds ($\lambda$) and reuse metrics ($\delta$).

\paragraph{Warmup Phase:} In contrast to prior caching methods, \foresight{} does not initialize the cache ($\mathcal{C}$) immediately after the first step. Instead, all spatial, temporal, cross-attention, and MLP blocks are computed for the first $W$ denoising steps, allowing intermediate features to stabilize. At timestep $t = W$, the cache is initialized with the latest outputs from each layer as described by Eq.~\ref{eq:cache_write}.

\begin{wrapfigure}{r}{0.65\linewidth}
\begin{center}
\vskip -0.1in
\centerline{\includegraphics[width=0.65\columnwidth]{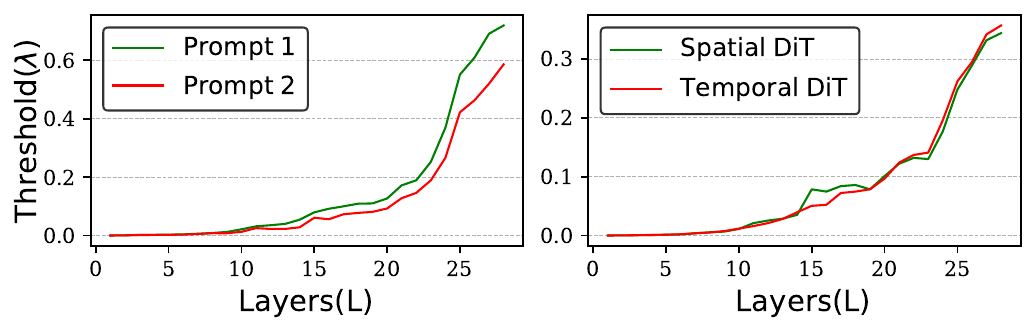}}
\caption{(\textbf{Left}) Spatial block thresholds for two different prompts using OpenSora (240p, 2s). (\textbf{Right}) Spatial and temporal block thresholds for the same prompt using OpenSora (720p, 2s).}
\label{fig:threshold}
\end{center}
\end{wrapfigure}

\foresight{} also establishes a reuse threshold ($\lambda$) for each layer based on the mean squared error (MSE) of intermediate features between consecutive steps during the warmup phase. Let $\mathbf{x}_t$ denote features at step $t$. Given stabilized features by $t=W$, we compute thresholds as a weighted sum of MSEs from the final three warmup steps, scaled geometrically to reduce bias:

\begin{equation} \label{eq:5}
    \lambda_{\mathbf{x}}^l = \sum_{t=W-2}^{W} \frac{1}{10^{W-t}} \left( \frac{1}{P} \sum_{i=1}^{P} \big( \mathbf{x}_{i}^l(t) - \mathbf{x}_{i}^l(t-1) \big)^2 \right),  
    \quad \mathbf{x} \in \{\mathbf{x}_\mathrm{spatial},\,\mathbf{x}_\mathrm{temporal}\}, 
    \quad  l = 1,\dots,L
\end{equation}

Here, $P = H\times W$ for spatial blocks and $P = F$ for temporal blocks. Figure~\ref{fig:threshold} shows the adaptive thresholds, showing variations for different prompts and resolutions. It shows how the reuse threshold varies if the resolution changes from 240p to 720p.

\paragraph{Reuse Phase:} The reuse phase employs the initialized cache ($\mathcal{C}$) and thresholds ($\lambda$), introducing a dynamic reuse metric ($\delta$) for each layer and block to guide reuse decisions. This phase alternates between reuse and recomputation steps. Specifically, reuse occurs for $N$ steps, after which all layers are recomputed for every $R$ step. The reuse metric is updated based on MSE between the current and cached features at each recomputation step, as defined in Eq~\ref{eq:6}. This update enables \foresight~to adapt and reuse dynamically based on feature changes.

\begin{equation} \label{eq:6}
    \delta_{\mathbf{x}}^l(t) = \frac{1}{P}\sum_{i=1}^{P}\big(\mathbf{x}_{i}^l(t)-\mathcal{C}_{i}^l(t-1)\big)^2,  
    \quad \mathbf{x} \in \{\mathbf{x}_\mathrm{spatial},\,\mathbf{x}_\mathrm{temporal}\}, 
    \quad l=1,\dots,L
\end{equation}

After updating the reuse metric, the cache is refreshed with the latest intermediate features defined in Eq.~\ref{eq:cache_write}. At the subsequent timestep ($t+1$), reuse decisions are made by comparing the reuse metric ($\delta$) against the threshold ($\lambda$):

\begin{equation} \label{eq:7}
\mathbf{x}_{t+1}^l =
\begin{cases}
    \mathcal{C}(\mathbf{x}_{t}^l), & \text{if} \quad \delta_{\mathbf{x}}^l(t)\leq \gamma\lambda_{\mathbf{x}}^l \\[4pt]
    \text{Compute}, & \text{otherwise}
\end{cases}, \quad \gamma\in(0,2],\; l=1,\dots,L
\end{equation}

We introduce a scaling factor ($\gamma$) to control the reuse threshold, enabling adjustable trade-offs between speed and quality: higher $\gamma$ emphasizes quality (less reuse), while lower $\gamma$ favors speed (more reuse).
Algorithm~\ref{alg:foresight} describes the end-to-end flow of the \foresight{} framework.

\input{./body/algorithms/algo_foresight2}

\begin{figure*}[t]
  \centering
  \includegraphics[width=1\columnwidth]{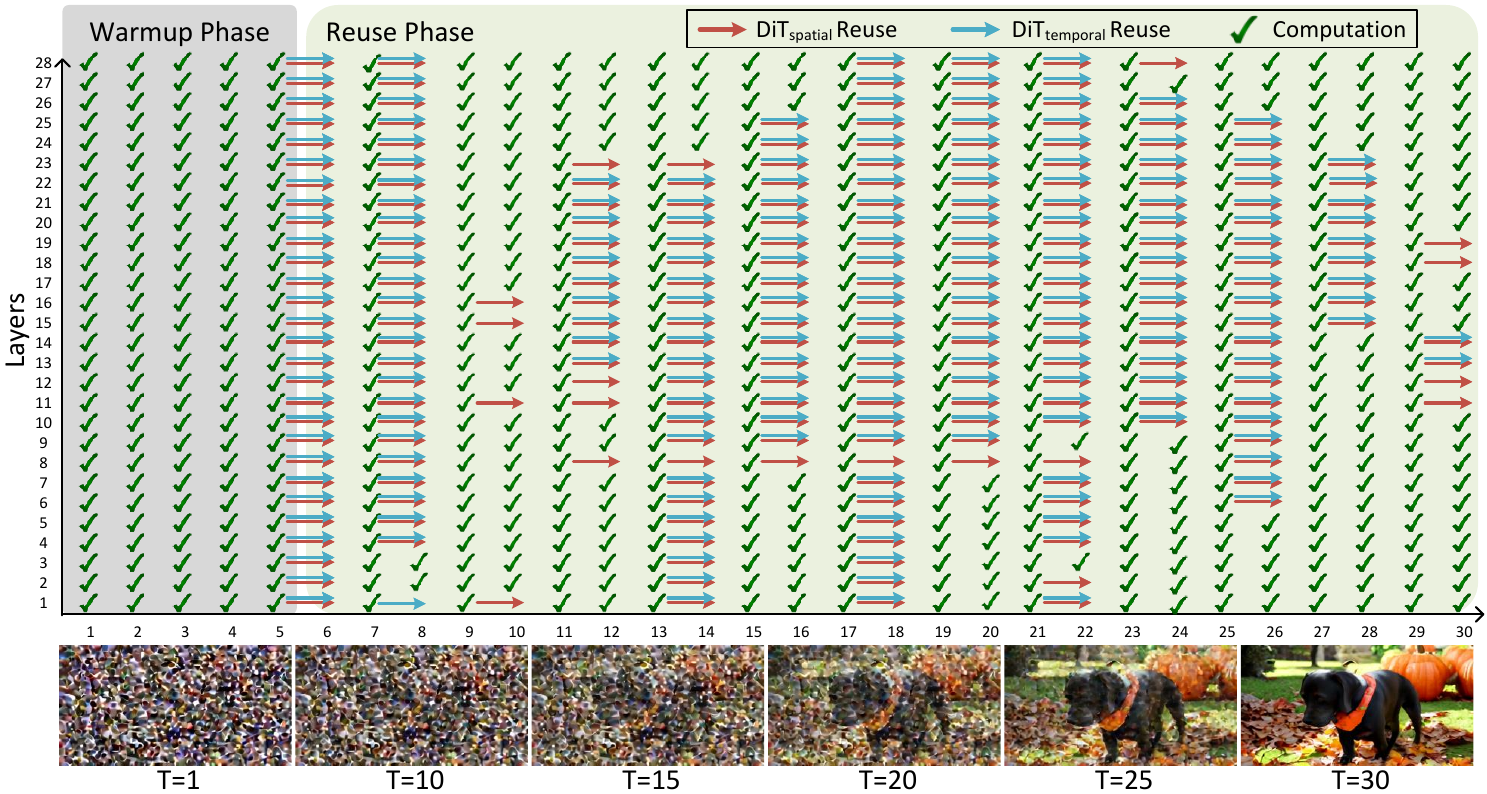}
  \caption{Example of \foresight's adaptive layer reuse on the OpenSora model (240p, 4s) with a warmup phase of $W=15\%$, reuse window of $N=1$, computation of $R=2$ and $\gamma=0.5$. The evolution of the video frame is shown at the bottom. The prompt is ``\emph{A playful black Labrador, adorned in a vibrant pumpkin-themed Halloween costume, surrounded by fallen leaves.}''
}
\label{fig:example}
\end{figure*}

\paragraph{Intuition with an Example:} Figure~\ref{fig:example} shows how \foresight{} works using OpenSora using an example prompt (as showcased in the caption). Checkmarks (\ding{52}) indicate computation, arrows ($\rightarrow$) represent reuse. During the warmup phase ($W=15\%$), all blocks are computed, initializing the cache and thresholds ($\lambda$) scaled by $\gamma=0.5$. In the reuse phase, a reuse window of $N=1$ and recomputation interval $R=2$ results in alternating reuse and recomputation steps. Notably, later layers (e.g., 24–28) are recomputed more frequently, demonstrating \foresight’s adaptive capability to maintain quality. \emph{The video frame evolution is shown at the bottom of Figure~\ref{fig:example}}.

\input{./body/convergence_analysis}

%% file: body/algorithms/algo_foresight2.tex
\begin{figure}[ht]
\vspace{-0.2in}
\centering
\resizebox{\linewidth}{!}{%
\begin{minipage}{1.23\linewidth}  
\begin{algorithm}[H]
\caption{\foresight}
\label{alg:foresight}

\newcommand{\funccommd}[1]{{\textcolor{blue}{#1}}}
\newcommand{\mycommfont}[1]{{\scriptsize\itshape\textcolor{darkgreen}{#1}}}

\begin{algorithmic}[1]
   \STATE {\bfseries Inputs:} Video tokens $\mathbf{x}_{H \times W}^{F}$, Steps $T$, Warmup steps $W$, Reuse steps $N$, Compute interval $R$
   \STATE {\bfseries Parameters:} Reuse threshold $\lambda$, scaling factor $\gamma$
   \STATE Initialize cache $\mathcal{C}_{\mathbf{x}}^l \gets \varnothing$

   \mycommfont{\# ===== Warmup Phase ===== \#}
   \FOR{$t=1$ {\bfseries to} $W$}
       \STATE $\mathbf{x}^l(t) \gets \funccommd{Compute}(\mathbf{x}^l), \quad \forall\, l=1,\dots,L$
       \STATE $\lambda_{\mathbf{x}}^l \gets \mathrm{MSE}[\mathbf{x}^l(t),\, \mathbf{x}^l(t{-}1)]$
   \ENDFOR
   \STATE $\mathcal{C}_{\mathbf{x}}^l \gets \mathbf{x}^l(W)$,\quad $\delta_{\mathbf{x}}^l \gets \lambda_{\mathbf{x}}^l$

   \mycommfont{\# ===== Reuse Phase ===== \#}
   \FOR{$t=W+1$ {\bfseries to} $T$}
       \IF{$t \mod R = 0$}
           \STATE $\mathbf{x}^l(t) \gets \funccommd{Compute}(\mathbf{x}^l),\quad \forall\, l=1,\dots,L$
           \STATE $\delta_{\mathbf{x}}^l \gets \mathrm{MSE}[\mathbf{x}^l(t),\, \mathcal{C}_{\mathbf{x}}^l]$
           \STATE $\mathcal{C}_{\mathbf{x}}^l \gets \mathbf{x}^l(t)$
       \ELSE
           \FOR{$l=1$ {\bfseries to} $L$}
               \IF{$\delta_{\mathbf{x}}^l \leq \gamma\lambda_{\mathbf{x}}^l$}
                   \STATE $\mathbf{x}^l(t) \gets \mathcal{C}_{\mathbf{x}}^l$
               \ELSE
                   \STATE $\mathbf{x}^l(t) \gets \funccommd{Compute}(\mathbf{x}^l)$
                   \STATE $\delta_{\mathbf{x}}^l \gets \mathrm{MSE}[\mathbf{x}^l(t),\, \mathcal{C}_{\mathbf{x}}^l]$
                   \STATE $\mathcal{C}_{\mathbf{x}}^l \gets \mathbf{x}^l(t)$
               \ENDIF
           \ENDFOR
       \ENDIF
   \ENDFOR
\end{algorithmic}
\end{algorithm}
\end{minipage}
}
\end{figure}

%% file: body/convergence_analysis.tex
\subsection{Foresight: Convergence Analysis}

We provide a formal analysis showing that \emph{adaptive reuse,} the key mechanism of \foresight, does not affect model convergence. Let $\mathbf{x}_0$ be the generated (denoised) video, $\mathbf{x}_t$ be the intermediate latent at denoising step $t$ ($T \leq t \leq 1$).  
For layer $l$ of the spatio-temporal diffusion transformer, denote its output by $f_{\theta}^{l}(\mathbf{x}_t)$. During timesteps $T_{\text{reuse}} \subseteq \{1, \dots, T\}$, adaptive reuse approximates this output using the cached value from the previous step as per Eq~\ref{eq:8}.  

\begin{equation} \label{eq:8}
\tilde{f}_{\theta}^{l}(\mathbf{x}_t) =
\begin{cases}
    f_{\theta}^{l}(\mathbf{x}_{t-1}) & \text{if } \text{MSE}(\mathbf{x}_t, \mathbf{x}_{t-1}) < \lambda_l \\
    f_{\theta}^{l}(\mathbf{x}_t) & \text{otherwise}
\end{cases}, \quad t=W+1,\dots,T,\; l=1,\dots,L
\end{equation}

Since reuse preserves weights $\theta$ and transition kernels $p_\theta(\mathbf{x}_{t-1} \mid \mathbf{x}_t)$, the original diffusion process is guaranteed to hold. Even when recomputed (in the worst case), the baseline forward pass is recovered. For any reused layer $l$ at denoising step $t$, the error is bounded as shown in Eq~\ref{eq:9}. 

\begin{equation} \label{eq:9}
   \| f_{\theta}^{l}(\mathbf{x}_t) - \tilde{f}_{\theta}^{l}(\mathbf{x}_t) \|_2 \leq \epsilon_l, \quad \epsilon_l \leq \sqrt{\gamma \lambda_l}
\end{equation}

Activations vary slowly in space/time when reused, making perturbations negligible. The reuse error at layer $l$ and step $t$ is shown in Eq~\ref{eq:10}.

\begin{equation} \label{eq:10}
   \varepsilon_t^l = \| f_{\theta}^{l}(\mathbf{x}_t) - \tilde{f}_{\theta}^{l}(\mathbf{x}_t) \|_2 \leq \underbrace{\sqrt{\gamma \lambda_x^l}}_{\varepsilon_{\text{max}}^l}
\end{equation}

Thus, $\varepsilon_t^l$ is uniformly bounded by a constant controllable via $\gamma$. Let $\mathbf{x}_t^*$ be the baseline latent and $\hat{\mathbf{x}}_t$ the latent under adaptive reuse. After one reverse step term 1 contracts by $\sqrt{1 - \beta_t} < 1$ and term 2 is bounded by $\varepsilon_{\text{tot}} = \sum_l L_l \varepsilon_{\text{max}}^l$ in Eq~\ref{eq:11}.

\begin{equation} \label{eq:11}
  \| \hat{\mathbf{x}}_{t-1} - \mathbf{x}_{t-1}^* \| \leq \sqrt{1 - \beta_t}  \| \hat{\mathbf{x}}_t - \mathbf{x}_t^* \| + \sum_{l=1}^L L_l \varepsilon_t^l
\end{equation}

Assuming each block is $L_l$-Lipschitz with $L_l < 1$ and $\varepsilon_t^l \leq \varepsilon_{\text{max}}^l$ for all $(t, l)$, unrolling the recursion for $k$ steps is shown in Eq~\ref{eq:12}.

\begin{equation} \label{eq:12}
     \| \hat{\mathbf{x}}_{t-k} - \mathbf{x}_{t-k}^* \| \leq \underbrace{\left( \prod_{s=t-k+1}^t \sqrt{1 - \beta_s} \right) \| \hat{\mathbf{x}}_t - \mathbf{x}_t^* \|}_{\text{(1)}} + \underbrace{\varepsilon_{\text{tot}} \sum_{j=0}^{k-1} \prod_{s=t-j+1}^t \sqrt{1 - \beta_s}}_{\text{(2)}}
\end{equation}

\begin{enumerate}
    \item As $k \to t$, $\prod_s \sqrt{1 - \beta_s} \to 0$ since $\sum_s \beta_s \to \infty$ implies $\prod_s (1 - \beta_s) \to 0$.
    \item Geometric series bounded by $\varepsilon_{\text{tot}} / (1 - \rho)$ where $\rho = \max_s \sqrt{1 - \beta_s} < 1$.
\end{enumerate}

As $t \to T$ and $T \to \infty$, (1) vanishes and (2) remains finite. Tightening $\gamma$ reduces $\varepsilon_{\text{tot}}$, yielding arbitrary closeness to the baseline. Adaptive reuse introduces bounded, vanishing perturbations into a contractive reverse Markov chain. Thus, \foresight maintains theoretical fidelity to the original diffusion model.

%% file: body/evaluation.tex
\section{Experiments}
\label{sec:experiments}

\subsection{Experimental Setup}
\label{subsec:exp_setup}

\paragraph{Models:} We evaluated \foresight{} on three popular text-to-video models: Open-Sora-v1.2~\citep{opensora}, Latte-1.0~\citep{latte}, and CogVideoX-2b~\citep{cogvideox}, using \textbf{NVIDIA A100} 80GB GPUs with FlashAttention~\citep{flashattention} enabled. Open-Sora used rflow~\citep{rlflow} sampling with 30 denoising steps and a CFG scale of 7.5. Latte and CogVideoX used DDIM~\citep{ddim} with 50 steps, and CFG scales of 7.5 and 6.0, respectively.

\paragraph{Evaluation Datasets and Metrics:} For comprehensive evaluation, we assess \foresight{} on the VBench benchmark suite~\citep{vbench}, using VBench accuracy, PSNR, SSIM~\citep{ssim}, LPIPS~\citep{lpips}, and FVD~\citep{fvd} to quantify video quality. We report results on the UCF-101~\citep{ucf101} and EvalCrafter~\citep{evalcrafter} prompt sets to evaluate performance on dynamic scenes. Following EvalCrafter, we also include $\mathrm{CLIPSIM}$, $\mathrm{CLIP\text{-}Temp}$, and DOVER’s VQA score~\citep{dover} for fine-grained quality assessment.

\paragraph{Baselines:} For a comprehensive comparison, we evaluate \foresight~against four training-free caching methods for accelerating diffusion models: $\mathrm{Static}$, $\Delta$-$\mathrm{DiT}$, $\mathrm{T\text{-}GATE}$, and $\mathrm{PAB}$. Implementation details for each baseline appear in Appendix~\ref{subsec:baseline_config}.

\subsection{Results}
\label{subsec:results}

\begin{table*}[t]
\centering
\input{./body/tables/results.tex}
\caption{Qualitative comparison of \foresight{} and static reuse methods on the VBench prompt set, which spans 11 dimensions with 50 prompts each. Videos are generated at 240p with OpenSora, 512x512 with Latte, and 480x720 with CogVideoX, all with a fixed duration of 2 seconds. Metrics including $\mathrm{PSNR}$, $\mathrm{SSIM}$, $\mathrm{LPIPS}$, and $\mathrm{FVD}$ are reported relative to the baseline.}
\label{tab:results}
\end{table*}

\paragraph{Quality Comparison:} Table~\ref{tab:results} compares the video quality of \foresight~with static reuse methods. We generate 550 videos from the VBench~\cite{vbench} prompt set, covering 550 prompts across 11 dimensions (50 per dimension). Video quality is evaluated using VBench accuracy (\%) and standard metrics: $\mathrm{PSNR}$, $\mathrm{SSIM}$, $\mathrm{LPIPS}$, and $\mathrm{FVD}$, all relative to the baseline. We report two \foresight~configurations, varying reuse steps from $N=1$ to $N=2$ and computation intervals from $R=2$ to $R=3$, with a fixed scaling factor ($\gamma=0.5$) balancing speed and quality. Additional results are provided in Appendix~\ref{subsec:additional_results}.

Qualitative results further highlight that \foresight{} surpasses static reuse methods by adaptively selecting which layers to reuse at each denoising step. This strategy preserves critical intermediate features, maintaining video quality and achieving up to \latencyimprv~speedup. \foresight{} consistently outperforms all baselines across varying noise schedulers, denoising steps, and architectures, demonstrating robustness and generalizability.

\paragraph{Memory Overhead:}

\begin{wrapfigure}{r}{0.7\linewidth}
\begin{center}
\vskip -0.2in
\centerline{\includegraphics[width=0.7\columnwidth]{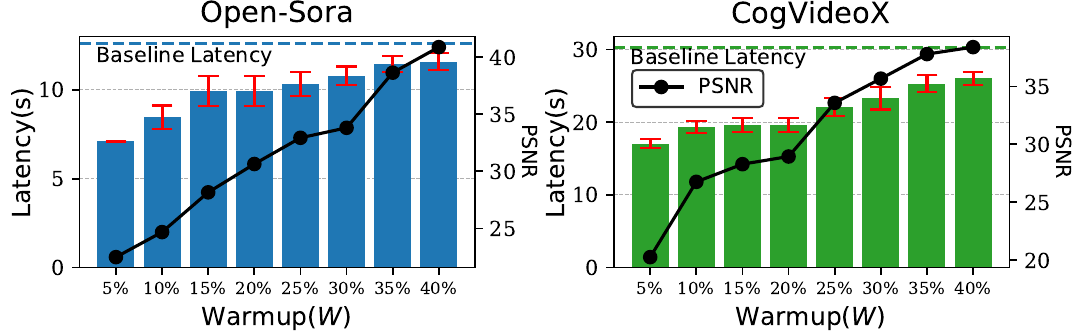}}
\caption{Varying warmup steps ($W$) with fixed reuse settings: $N=1$, $R=2$, and scaling factor $\gamma=0.5$ in \foresight.}
\label{fig:warmup}
\end{center}
\vskip -0.2in
\end{wrapfigure}

While caching techniques require no training, they introduce memory overhead from storing intermediate features. This overhead grows with model depth, resolution, and video length. Prior work~\citep{pab} employs fine-grained caching across spatial, temporal, cross-attention, and MLP blocks in all layers, resulting in a cache size of $\mathcal{C} = 6LHWF$, where $L$ is the number of layers, $F$ the number of frames, and each frame has spatial size $H \times W$. In contrast, \foresight{} adopts a coarser strategy, caching only full DiT blocks (two per layer), reducing the cost to $\mathcal{C} = 2LHWF$. Thus, \foresight{} provides a \textbf{3$\times$} higher memory reduction.

\subsection{Ablation Studies}

We conduct ablation studies to analyze \foresight's behavior by varying its key hyperparameters and examining the trade-offs between speed and video quality.

\paragraph{Varying Warmup Phase $(W)$:}

To study the impact of the warmup phase, we vary its length from 5\% to 40\%, keeping all other \foresight settings fixed. As shown in Figure~\ref{fig:warmup}, increasing the warmup percentage reduces the number of denoising steps available for reuse. This leads to higher video quality but limits performance gains, as reuse is applied to fewer steps.

\input{./body/tables/ablation2}

\paragraph{Varying Reuse Settings $(N,R)$:} 
Reuse settings determine the maximum reuse across adjacent timesteps. We conducted an ablation study to analyze their impact by varying the reuse steps and the compute interval.
Table~\ref{tab:reuse_settings} summarizes the inference latency and video quality trade-off of \foresight{} relative to PAB as we vary the reuse window ($N$) and compute interval ($R$). Larger $N$ increases reuse opportunities but lowers the frequency of cache and reuse metric updates, eventually harming fidelity. In practice, we observe that for the range of $N{=}1$ to $N{=}3$ delivers the best balance, after which the video quality drops compared to the PAB baseline, mainly due to not updating the reuse metric. For $N{=}3$, \foresight{} yields a 1.35$\times$ speed-up while surpassing PAB in video quality.

\paragraph{Varying Scaling Factor $(\gamma)$:} The scaling factor $\gamma$ governs the speed–quality balance by adjusting how strictly reuse is applied. The reuse threshold is dynamic and set at the end of the warmup phase. The scaling factor can be applied uniformly across all layers or adjusted per layer to account for varying sensitivity.
Table~\ref{tab:scale_factor} shows that smaller values increase latency but enhance fidelity.  With $\gamma{=}0.25$, \foresight{} reaches a $\mathrm{PSNR}$ of 39, which is 9.97 points above PAB, at the cost of just a 0.62\,s latency increase. This reflects a slight reduction in the reuse profile.

%% file: body/tables/results.tex
\resizebox{1\textwidth}{!}{
\begin{tabular}{l c c c c c c c c}
\toprule
\textbf{Model} & \textbf{Method} &
$\mathrm{VBench (\%)}\uparrow$ & $\mathrm{PSNR}\uparrow$ &  $\mathrm{SSIM}\uparrow$ & $\mathrm{LPIPS}\downarrow$ & $\mathrm{FVD}\downarrow$ & $\mathrm{Latency(s)}$ & $\mathrm{Speedup}\uparrow$
\\
\toprule
 \multirow{7}{*}{\textbf{Open-Sora}} & $\mathrm{Baseline}$ & 75.63 & - &	- &	- &	- &	12.65 ($\pm\;$0.06) & -
 \\
 & $\mathrm{Static}$ & 73.10 & 20.25 & 0.71 &	0.21 & 1257.50 &	9.27 ($\pm\;$0.03) &	1.36$\times$
 \\
 & $\Delta$-$\mathrm{DiT}$ & 74.20 &	22.45 & 0.73 &	0.18 & 1009.65 &	11.95 ($\pm\;$0.03) &	1.06$\times$
 \\
 & $\mathrm{T\text{-}GATE}$ & 73.80 &	21.78 & 0.71 &	0.20 & 890.45 &	11.1 ($\pm\;$0.02) & 1.14$\times$ \\
 & $\mathrm{PAB}$ & 75.32 &	25.67 & 0.85 &	0.10 &	541.53 &	10.01 ($\pm\;$0.03) &	1.26$\times$
\\
 & \foresight~($N_1R_2$) & \textbf{75.90} & \textbf{29.67} & \textbf{0.90} & \textbf{0.05} &	\textbf{306.66} & 9.92 ($\pm\;$0.92) & 1.28$\times$
 \\
 & \foresight~($N_2R_3$) & 75.62 & 27.49 &	0.87 &	0.07 &	457.69 &	\textbf{8.80 ($\pm\;$1.12)} &	\textbf{1.44$\times$}
 \\
 \midrule
 \multirow{7}{*}{\textbf{Latte}} & $\mathrm{Baseline}$ & 73.02 &	- &	- &	- &	- &	33.53 ($\pm\;$0.09) &	-
 \\
 & $\mathrm{Static}$ & 61.56 & 10.41 &	0.36 &	0.65 &	3937.29 &	25.90 ($\pm\;$0.09)	&	1.29$\times$
\\
 & $\Delta$-$\mathrm{DiT}$ & 65.14 &	14.85	& 0.51 &	0.48 &	2751.64 &	32.84 ($\pm\;$0.02) &	1.02$\times$
 \\
 & $\mathrm{T\text{-}GATE}$ & 62.48 &	12.56 &	0.48 &	0.56 &	1894.45 &	30.54 ($\pm\;$0.05) &	1.09$\times$
\\
 & $\mathrm{PAB}$ & 72.64 &	21.22 &	0.75 &	0.23 &	882.96 &	25.93 ($\pm\;$0.02) &	1.29$\times$
\\
 & \foresight~($N_1R_2$) & \textbf{73.08} &	\textbf{26.02} &	\textbf{0.86} &	\textbf{0.12} &	\textbf{443.05} &	28.40 ($\pm\;$1.36) & 1.18$\times$
 \\
 & \foresight~($N_2R_3$) & 73.05 &	23.12 &	0.82 &	0.18 &	683.18 &	\textbf{25.54 ($\pm\;$1.56)} &	\textbf{1.31$\times$}
\\
 \midrule
 \multirow{7}{*}{\textbf{CogVideoX}} & $\mathrm{Baseline}$ & 77.78 & - &	- &	- &	- &	30.23 ($\pm\;$0.03) &	-
  \\
 & $\mathrm{Static}$ & 76.65 &	14.05 &	0.63 &	0.42 &	1811.77 &	22.44 ($\pm\;$0.01) &	1.35$\times$
\\
 & $\Delta$-$\mathrm{DiT}$ & 54.16 &	11.94 &	0.49 &	0.65 &	3085.41 &	29.12 ($\pm\;$0.01) &	1.03$\times$
\\
 & $\mathrm{T\text{-}GATE}$ & 65.14 &	12.56 &	0.45 &	0.64 &	2845.64 &	27.64 ($\pm\;$0.03) &	1.09$\times$
\\
 & $\mathrm{PAB}$ & 77.89 &	29.04 &	0.91 &	0.07 &	340.24 &	22.05 ($\pm\;$0.01) &	1.37$\times$
\\
 & \foresight~($N_1R_2$) & \textbf{77.94} &	\textbf{34.75} &	\textbf{0.95} &	\textbf{0.03} &	\textbf{130.65} &	20.67 ($\pm\;$1.10) &	1.46$\times$
 \\
 & \foresight~($N_2R_3$) & 77.84 &	28.45 &	0.87 &	0.12 &	531.99 &	\textbf{18.53 ($\pm\;$1.49)} &	\textbf{1.63$\times$}
\\
 \bottomrule
\end{tabular}}

%% file: body/tables/ablation2.tex
\begin{table}[t]
\centering
\setlength{\tabcolsep}{5pt}  
\renewcommand{\arraystretch}{1.2}  

\begin{minipage}{0.48\textwidth}
\fbox{%
\begin{minipage}{0.95\linewidth}
\centering
\caption{Effect of reuse settings on Open-Sora (240p, 2s, $T{=}60$, $W{=}15\%$, $\gamma{=}0.5$). Compared to PAB ($\mathrm{Latency}=19.88$s, $\mathrm{PSNR}=28.12$).}
\label{tab:reuse_settings}
\vspace{4pt}
\begin{tabular}{@{}lcc@{}}
\toprule
Settings & Latency $\downarrow$ & PSNR $\uparrow$ \\
\midrule
$N{=}1,\;R{=}2$ & 18.70 {\scriptsize\textcolor{darkgreen}{(–1.17)}} & 32.38 {\scriptsize\textcolor{darkgreen}{(+4.26)}} \\
$N{=}2,\;R{=}3$ & 16.37 {\scriptsize\textcolor{darkgreen}{(–3.50)}} & 30.42 {\scriptsize\textcolor{darkgreen}{(+2.30)}} \\
$N{=}3,\;R{=}4$ & 14.79 {\scriptsize\textcolor{darkgreen}{(–5.08)}} & 29.03 {\scriptsize\textcolor{darkgreen}{(+0.91)}} \\
$N{=}4,\;R{=}5$ & 13.49 {\scriptsize\textcolor{darkgreen}{(–6.38)}} & 27.91 {\scriptsize\textcolor{red}{(–0.20)}} \\
\bottomrule
\end{tabular}
\vspace{2pt}
\end{minipage}
}
\end{minipage}
\hfill
\begin{minipage}{0.48\textwidth}
\fbox{%
\begin{minipage}{0.95\linewidth}
\centering
\caption{Effect of scaling factor on Open-Sora ($N{=}1$, $R{=}2$, 240p, 2s, $T{=}60$, $W{=}15\%$). Compared to PAB ($\mathrm{Latency}=19.88$s, $\mathrm{PSNR}=28.12$).}
\label{tab:scale_factor}
\vspace{4pt}
\begin{tabular}{@{}lcc@{}}
\toprule
$\gamma$ & Latency $\downarrow$ & PSNR $\uparrow$ \\
\midrule
0.25 & 20.50 {\scriptsize\textcolor{red}{(+0.62)}}  & 38.09 {\scriptsize\textcolor{darkgreen}{(+9.97)}} \\
0.5  & 18.70 {\scriptsize\textcolor{darkgreen}{(–1.17)}} & 32.38 {\scriptsize\textcolor{darkgreen}{(+4.26)}} \\
1.0  & 17.03 {\scriptsize\textcolor{darkgreen}{(–2.84)}} & 30.43 {\scriptsize\textcolor{darkgreen}{(+2.31)}} \\
2.0  & 16.02 {\scriptsize\textcolor{darkgreen}{(–3.85)}} & 29.51 {\scriptsize\textcolor{darkgreen}{(+1.39)}} \\
\bottomrule
\end{tabular}
\vspace{2pt}
\end{minipage}
}
\end{minipage}
\end{table}

%% file: body/limitations.tex
\section{Limitations and Discussion}
\label{sec:limitations}

\foresight~operates at runtime without altering model architecture or application. Its plug-and-play design enables efficient deployment of text-to-video models, reducing computational costs and expanding access to generative AI for creative and educational purposes. However, broader accessibility may also increase the risk of misuse, such as creating deepfakes for misinformation or fraud. While \foresight~does not inherently heighten these risks, its widespread use underscores the need for safeguards and ethical guidelines with controlled use of the model and usage guidelines.
\foresight{} accelerates inference by reusing intermediate activations; its speed and video quality envelope are governed by the reuse window $N$ and warm-up length $W$. Essentially, a larger $N$ or a smaller $W$ could yield greater speed-up at the risk of quality loss. Furthermore, \foresight{} can be expanded. Our experiments adopt coarse, block-level caching, yet the framework can operate at finer granularity.

%% file: body/conclusion.tex
\section{Conclusions}
\label{sec:conclusion}
We introduced \textbf{\foresight}, a training-free framework that accelerates diffusion transformers for text-to-video generation by making layer-wise reuse decisions at run time. Rather than relying on static caching, \foresight{} dynamically tracks mean-squared-error (MSE) feature change, maintains per-layer reuse thresholds, and selectively decides whether to reuse or recompute each block. Evaluated on Open-Sora-v1.2, Latte-1.0, and CogVideoX-2b, \foresight{} achieves up to \latencyimprv speed-up in inference latency while consistently improving PSNR, SSIM, LPIPS, and FVD compared to state-of-the-art baselines. Its coarse, block-level caching strategy also reduces memory usage by 3$\times$, without requiring retraining or architectural modifications.

\section*{Acknowledgments}
Muhammad Adnan’s Ph.D. is supported by the Intel TSA
and Natural Sciences and Engineering Research Council of Canada (NSERC) [RGPIN-2019-05059] Grants.

%% file: body/appendix.tex
\section*{Appendix}

\input{./body/background}

\section{Workload Characterization}
\label{subsec:workloadcharacterization}

To analyze inference bottlenecks in text-to-video generation, we profile the Open-Sora~\citep{opensora} model across various resolutions and timeframes, using a fixed 30-step RFlow scheduler using a single NVIDIA A100 (80GB) GPU with flash attention~\citep{flashattention}.

\begin{figure}[t]
  \centering
  \includegraphics[width=0.7\textwidth]{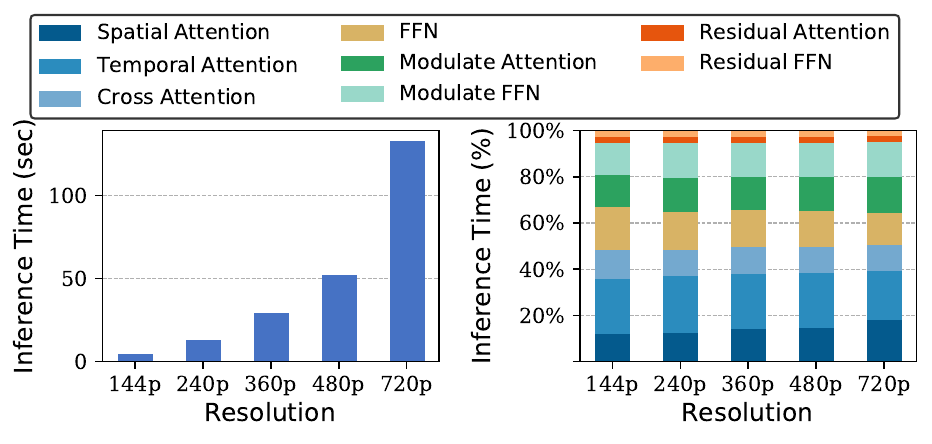}
  \caption{\textbf{Left:} End-to-end inference time across resolutions. \textbf{Right:} Inference time breakdown by operator. Results based on Open-Sora~\citep{opensora} using a single NVIDIA A100 (80GB) GPU with batch size 1.
}
\label{fig:charaterization}
\end{figure}

Figure~\ref{fig:charaterization} shows the end-to-end latency and its breakdown by operator type. As resolution increases from 480p to 720p, latency rises 2.5$\times$, driven largely by the quadratic complexity of attention operations.
Attention modules (spatial, temporal, and cross-attention) account for ~50\% of inference time, FFNs for 15\%, and non-linear layers—LayerNorm, scaling, and residuals—for 35\%. The sizable cost of non-linear operations highlights the need to target them in optimization efforts.

\begin{figure}[h]
  \centering
  \includegraphics[width=0.7\textwidth]{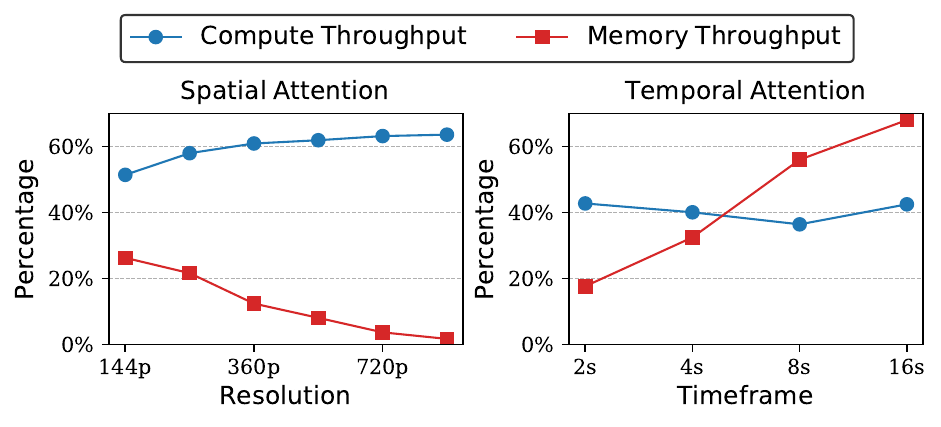}
  \caption{Compute vs. memory throughput for Spatial and Temporal Attention blocks in Open-Sora~\citep{opensora}, measured on a single NVIDIA A100 (80GB) GPU with batch size 1.
}
\label{fig:bottleneck}
\end{figure}

To identify system bottlenecks during inference, we measured compute and memory throughput for Spatial and Temporal Attention blocks in the Open-Sora~\citep{opensora} model using a single NVIDIA A100 (80GB) GPU with batch size 1 (Figure~\ref{fig:bottleneck}). For Spatial Attention, we varied resolution from 144p to 1080p with a fixed timeframe of 8 seconds. For Temporal Attention, we varied timeframes from 2 to 16 seconds at a fixed resolution of 720p.

In Spatial Attention, increasing resolution increases the number of spatial tokens, leading to higher compute demands due to the quadratic complexity of attention. This keeps the block compute-bound. In contrast, Temporal Attention sees a smaller increase in sequence length with longer timeframes, while the batch size—equal to the number of token patches in a frame—remains large. As a result, flash attention becomes suboptimal and performance becomes memory-bound. Overall, ST-DiT layers are primarily compute-bound, suggesting that reusing redundant computations can improve inference efficiency.

\begin{figure*}[t!]
  \centering
  \includegraphics[width=1\columnwidth]{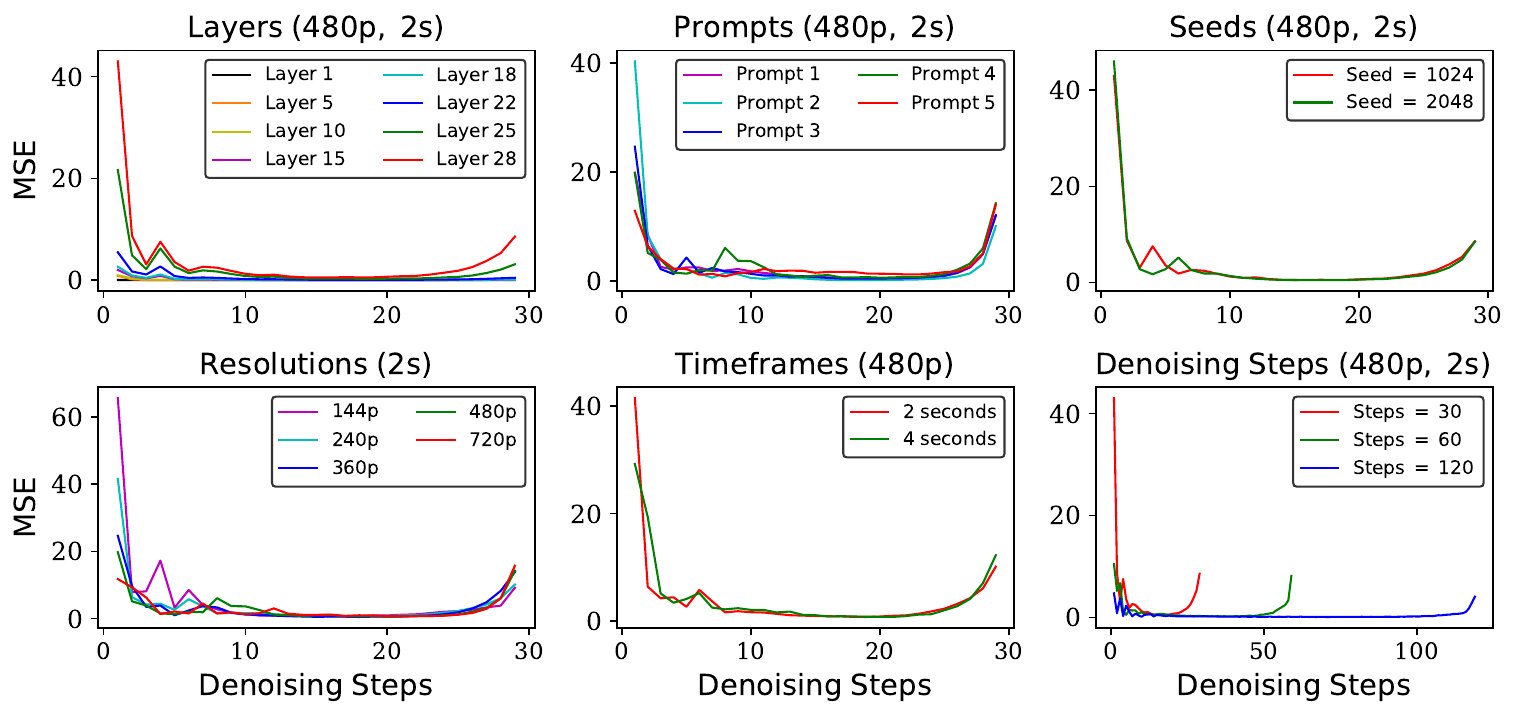}
  \caption{Quantitave analysis of Spatial DiT output using mean squared difference across different layer, prompts, seeds, video resolutions, video timeframes and denoising steps for OpenSora~\cite{opensora} model Layer 28 if layer not specified. The prompt is `` \emph{A narrow, cobblestone alleyway, bathed in the soft glow of vintage street lamps, stretches between tall, weathered brick buildings adorned with ivy. The scene begins with a gentle drizzle, creating a reflective sheen on the cobblestones. As the camera pans, a black cat with piercing green eyes darts across the path, adding a touch of mystery. The alley is lined with quaint, shuttered windows and wooden doors, some slightly ajar, hinting at hidden stories within. A soft breeze rustles the leaves of potted plants and hanging flower baskets, while distant, muffled sounds of city life create a serene yet vibrant atmosphere.}'' 
}
\label{fig:mse}
\end{figure*}

\section{Feature Variations and Reuse Metric}
\label{subsec:videoconfigs}

To analyze feature variation across video configurations, we vary one parameter at a time while keeping others fixed, and measure changes in intermediate features at specific layers. Figure~\ref{fig:mse} shows the effect of varying prompts, noise seeds, resolution, timeframes, and denoising steps. The results indicate that intermediate features are sensitive to these configurations. Therefore, adaptive reuse must account for such variations to minimize quality loss in video generation.

\subsection{Reuse Metric}
\label{subsec:reuse_metric}

To analyze the dynamic behavior of reuse and intermediate feature variation, we measure how features evolve across layers and timesteps using cosine similarity.

\subsubsection{Across Condition}

\begin{figure}[h!]
  \centering
  \includegraphics[width=0.5\textwidth]{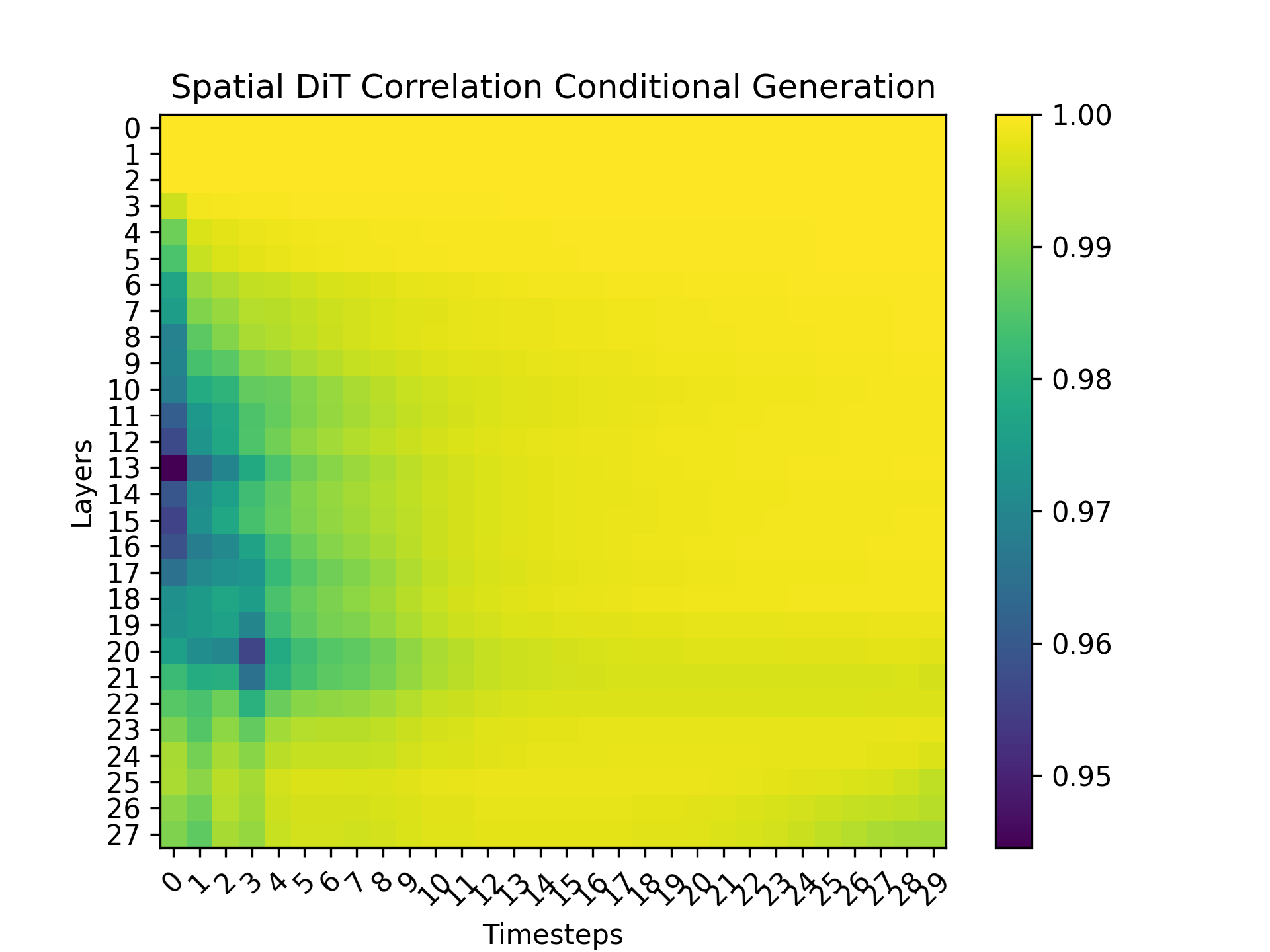}
  \caption{Cosine similarity of Spatial-DiT features across conditioning steps in the OpenSora model.
}
\label{fig:condition}
\end{figure}

Figure~\ref{fig:condition} illustrates the evolution of spatial features during conditional generation, across layers and denoising steps.

\subsubsection{Across Layers}

\begin{figure*}[t]
  \centering
  \subfloat[$\mathrm{T=5}$]{
  \begin{minipage}[t]{0.48\textwidth}
      \centering
      \includegraphics[width=\textwidth]{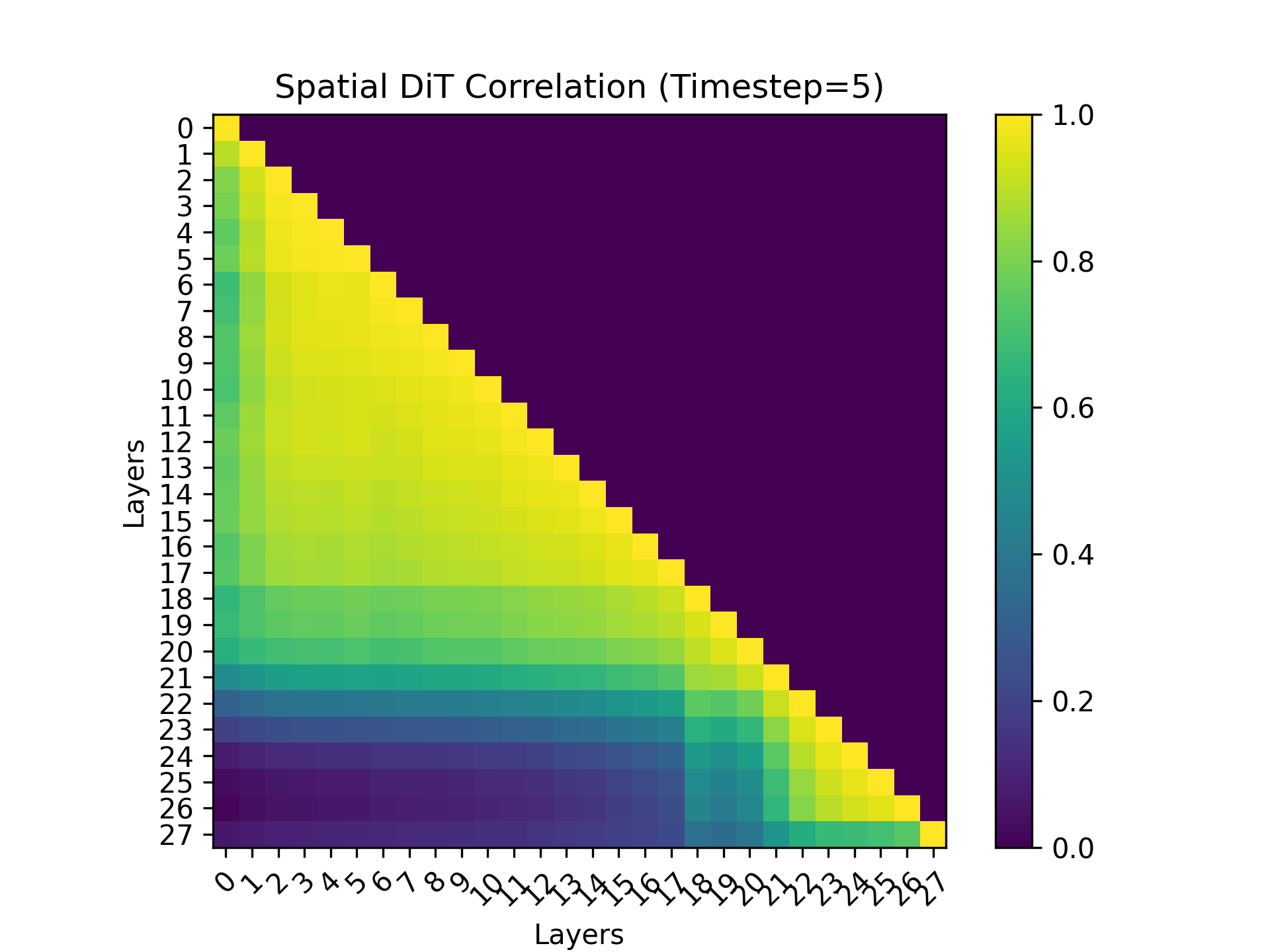}
  \end{minipage}
  }
  \hfill
  \subfloat[$\mathrm{T=10}$]{
  \begin{minipage}[t]{0.48\textwidth}
  \centering
  \includegraphics[width=\textwidth]{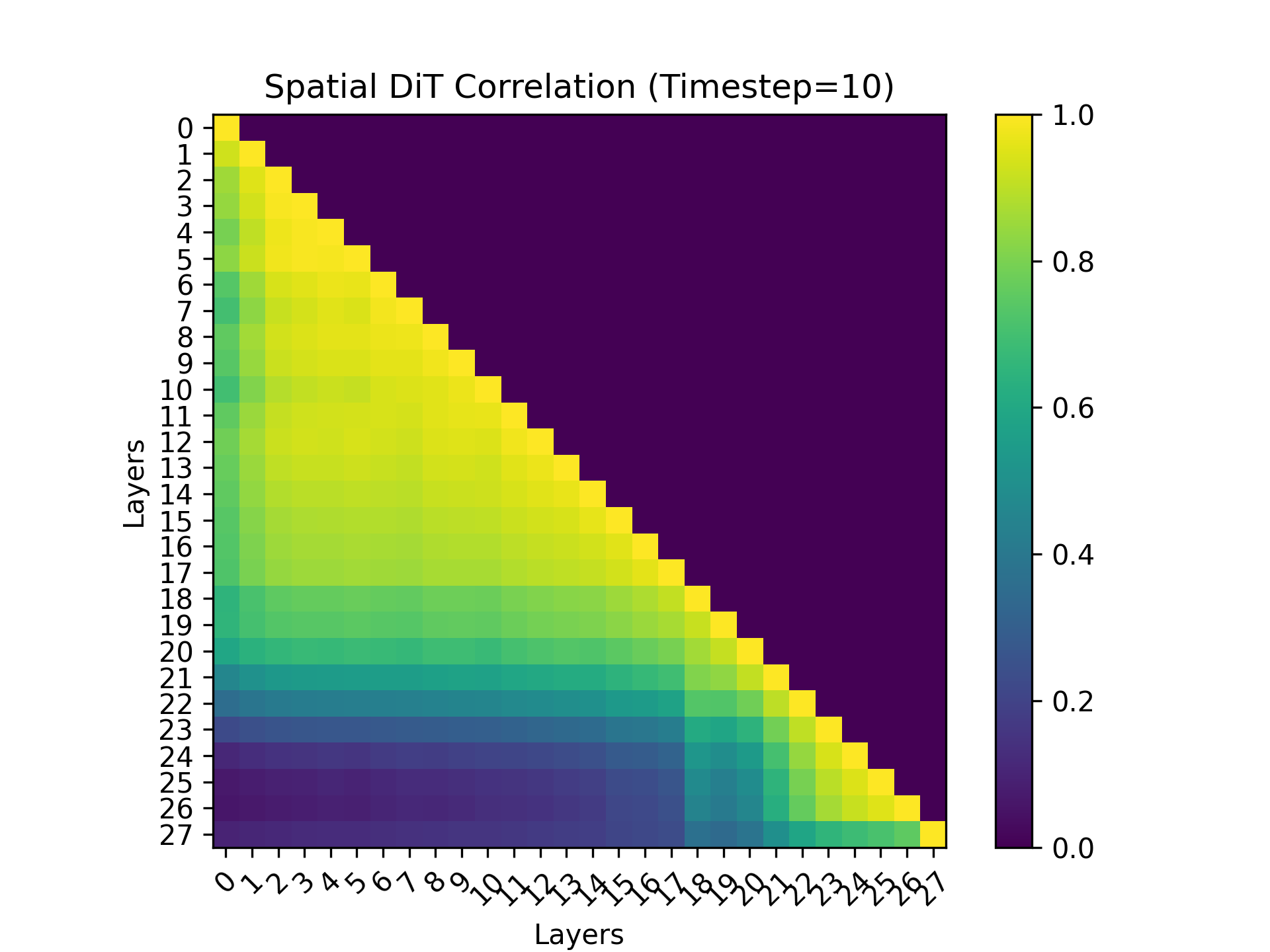}
  \end{minipage}
  }
  
  
  \subfloat[$\mathrm{T=15}$]{
  \begin{minipage}[t]{0.48\textwidth}
      \centering
      \includegraphics[width=\textwidth]{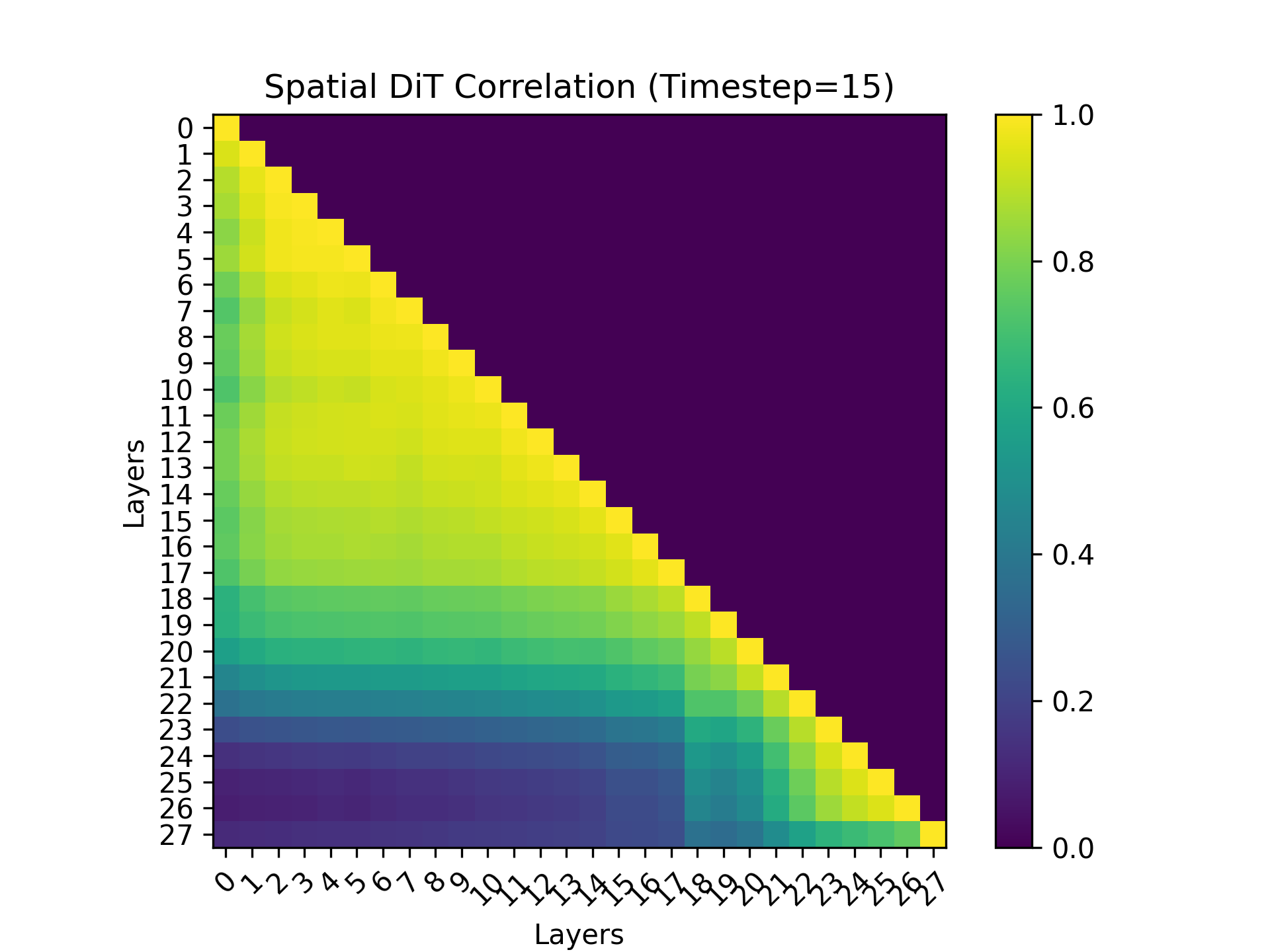}
  \end{minipage}
  }
  \hfill
  \subfloat[$\mathrm{T=20}$]{
  \begin{minipage}[t]{0.48\textwidth}
  \centering
  \includegraphics[width=\textwidth]{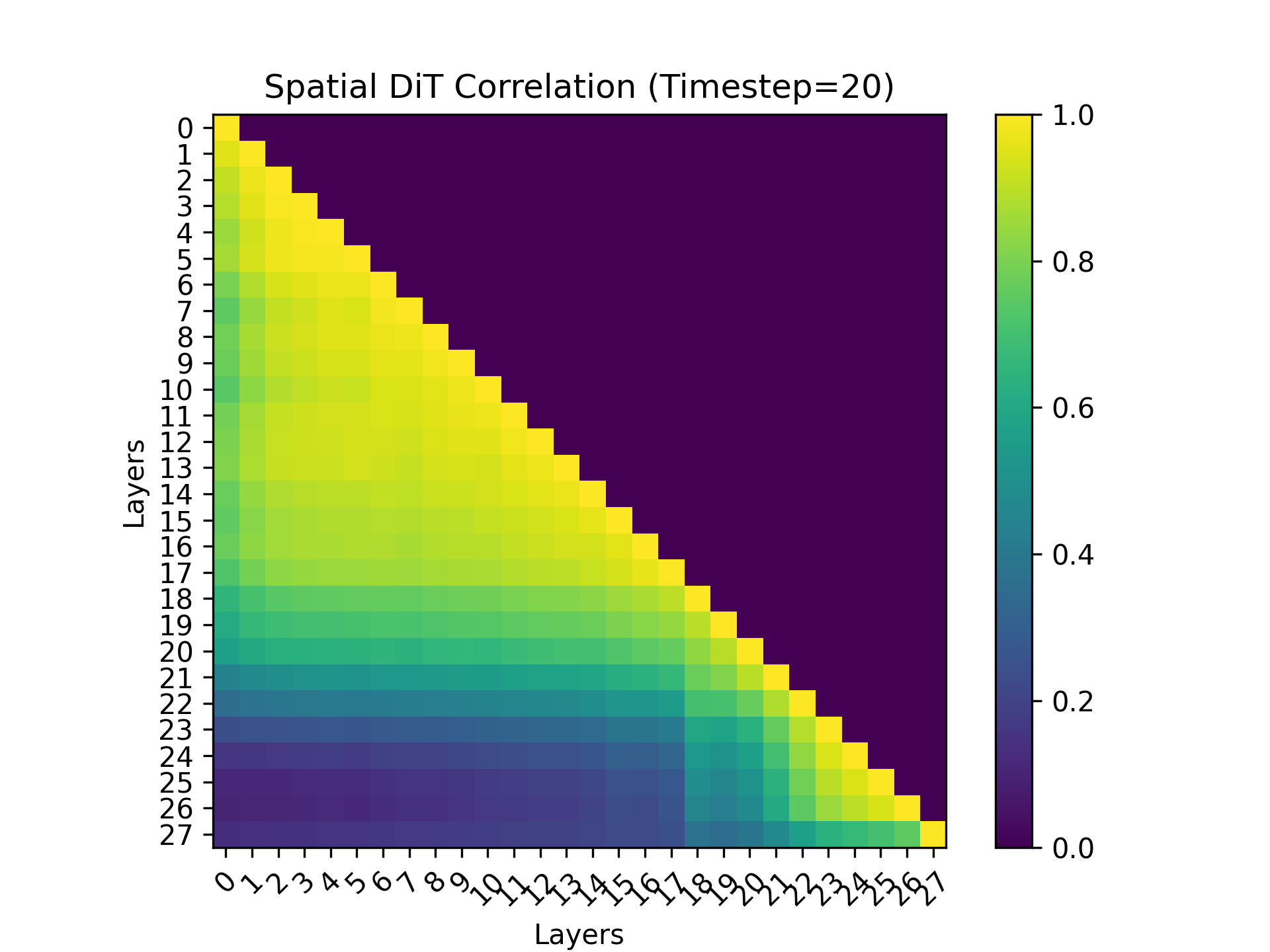}
  \end{minipage}
  }


\subfloat[$\mathrm{T=25}$]{
  \begin{minipage}[t]{0.48\textwidth}
      \centering
      \includegraphics[width=\textwidth]{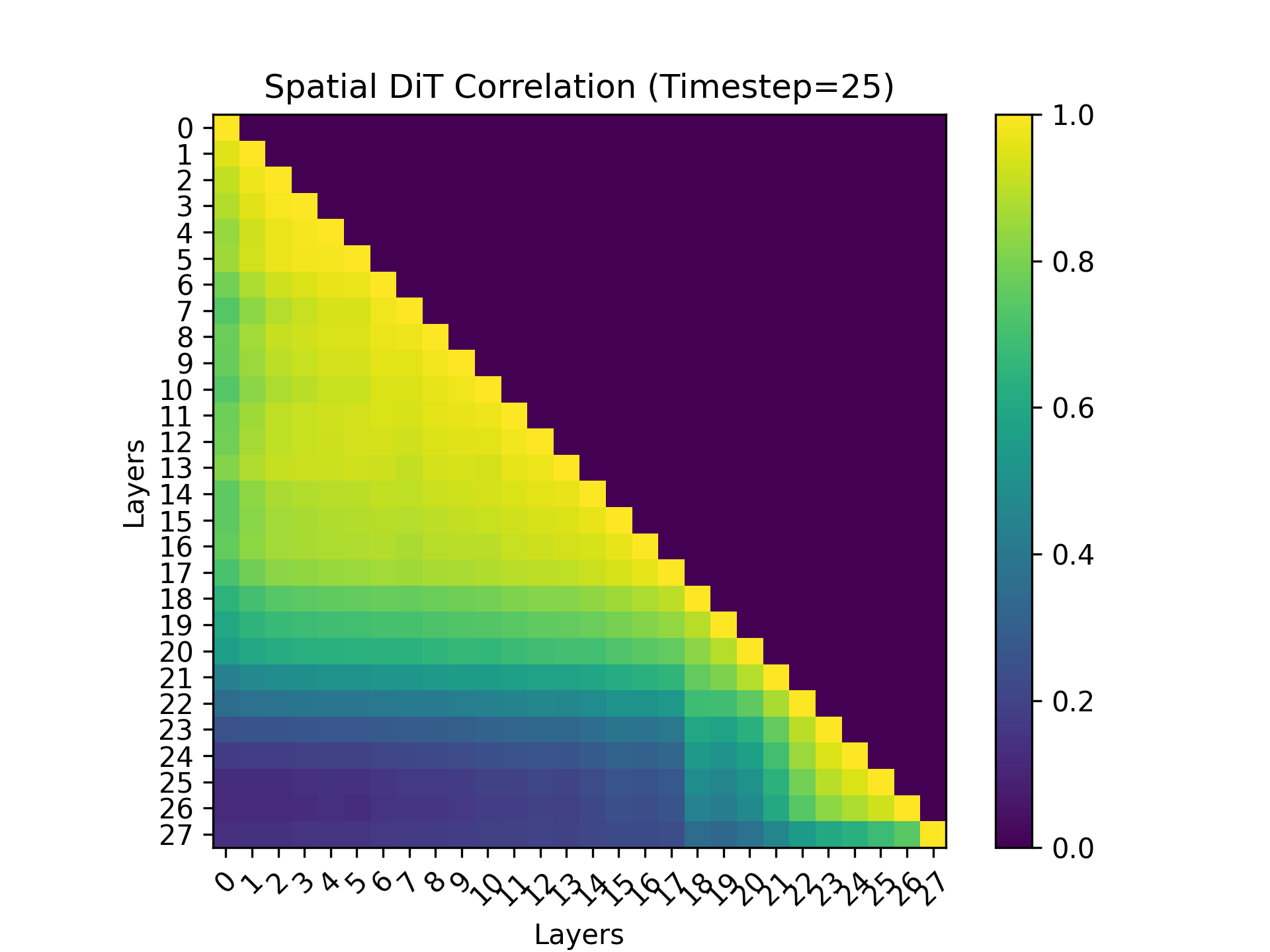}
  \end{minipage}
  }
  \hfill
  \subfloat[$\mathrm{T=30}$]{
  \begin{minipage}[t]{0.48\textwidth}
  \centering
  \includegraphics[width=\textwidth]{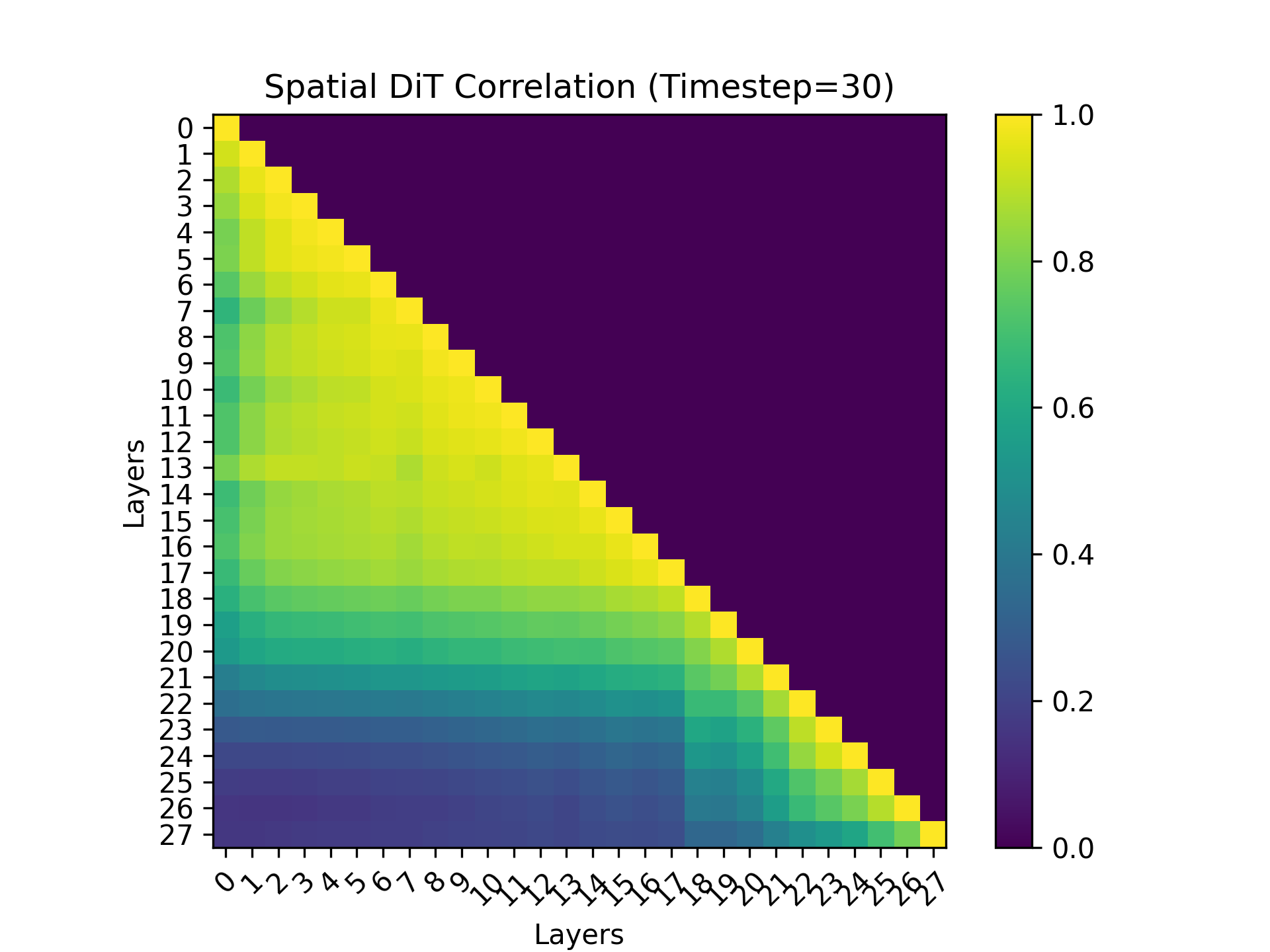}
  \end{minipage}
  }
  
  \caption{Cosine similarity of Spatial-DiT features across layers at different denoising steps in the OpenSora model.}
  \label{fig:across_layers}
\end{figure*}

Figure~\ref{fig:across_layers} shows the cosine similarity of spatial-DiT features across layers and denoising steps for the OpenSora model.

\subsubsection{Across Denoising Steps}

\begin{figure*}[t]
  \centering
  \subfloat[$\mathrm{Layer\;4}$]{
  \begin{minipage}[t]{0.48\textwidth}
      \centering
      \includegraphics[width=\textwidth]{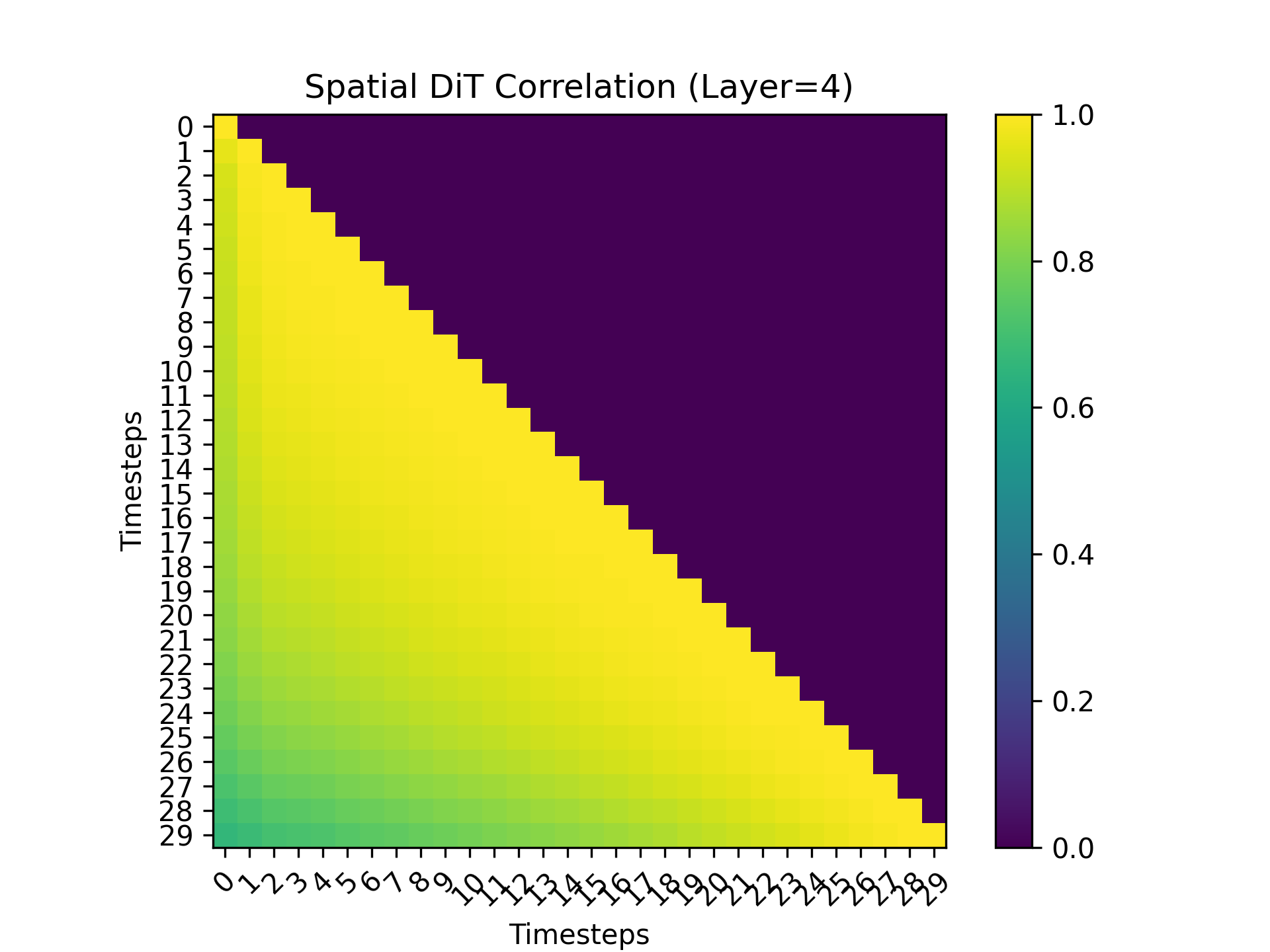}
  \end{minipage}
  }
  \hfill
  \subfloat[$\mathrm{Layer\;9}$]{
  \begin{minipage}[t]{0.48\textwidth}
  \centering
  \includegraphics[width=\textwidth]{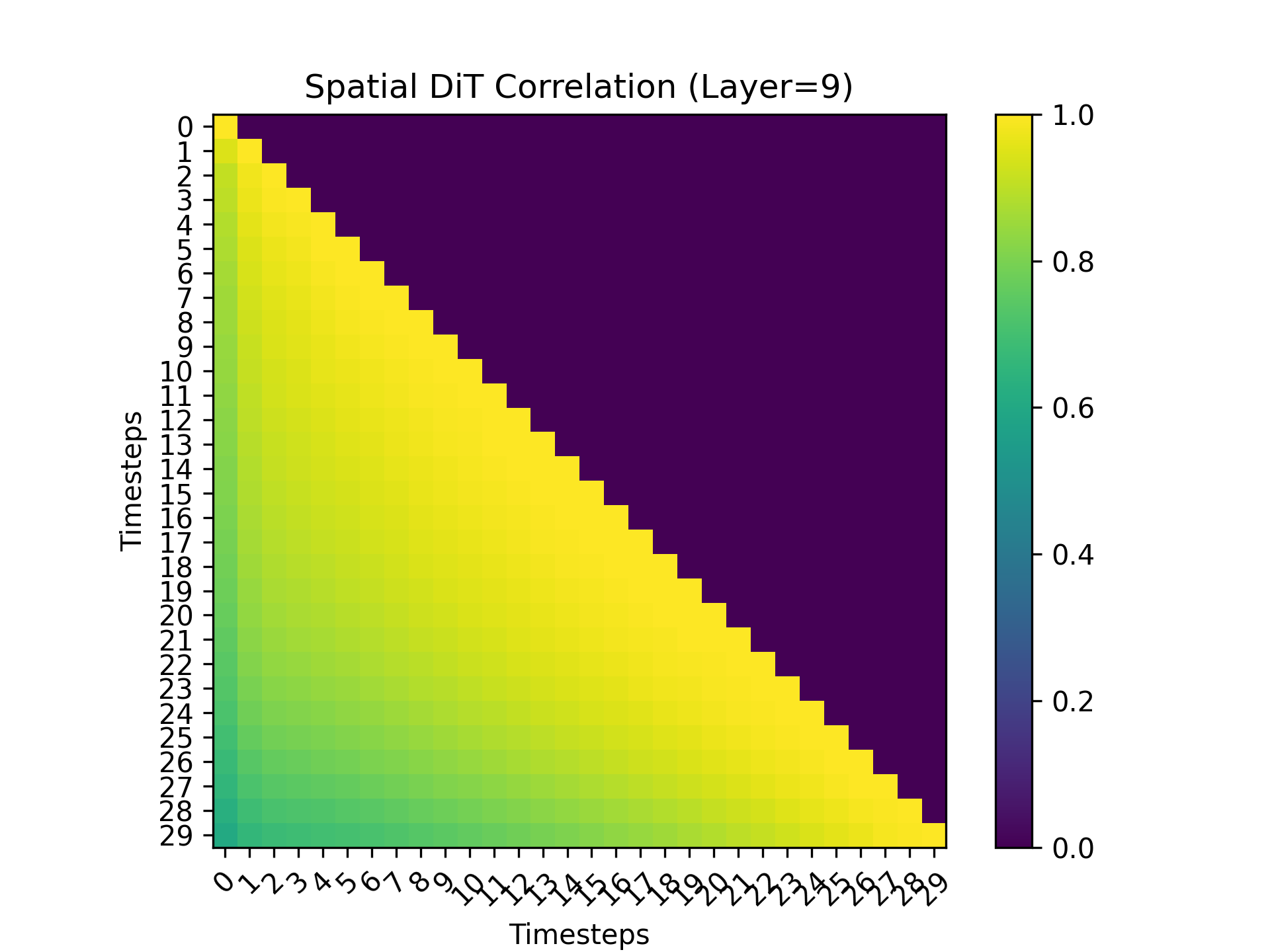}
  \end{minipage}
  }
  
  
  \subfloat[$\mathrm{Layer\;14}$]{
  \begin{minipage}[t]{0.48\textwidth}
      \centering
      \includegraphics[width=\textwidth]{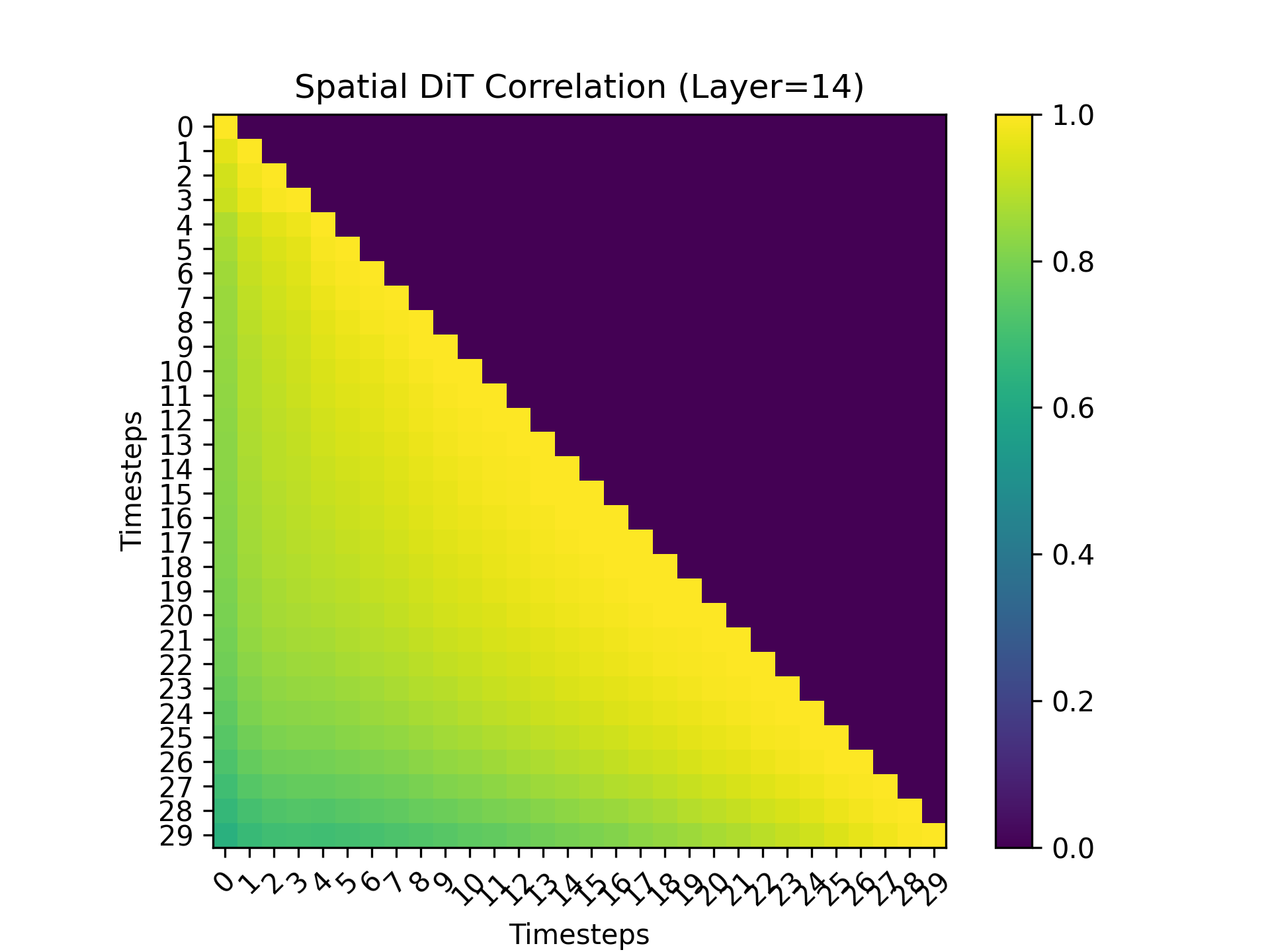}
  \end{minipage}
  }
  \hfill
  \subfloat[$\mathrm{Layer\;19}$]{
  \begin{minipage}[t]{0.48\textwidth}
  \centering
  \includegraphics[width=\textwidth]{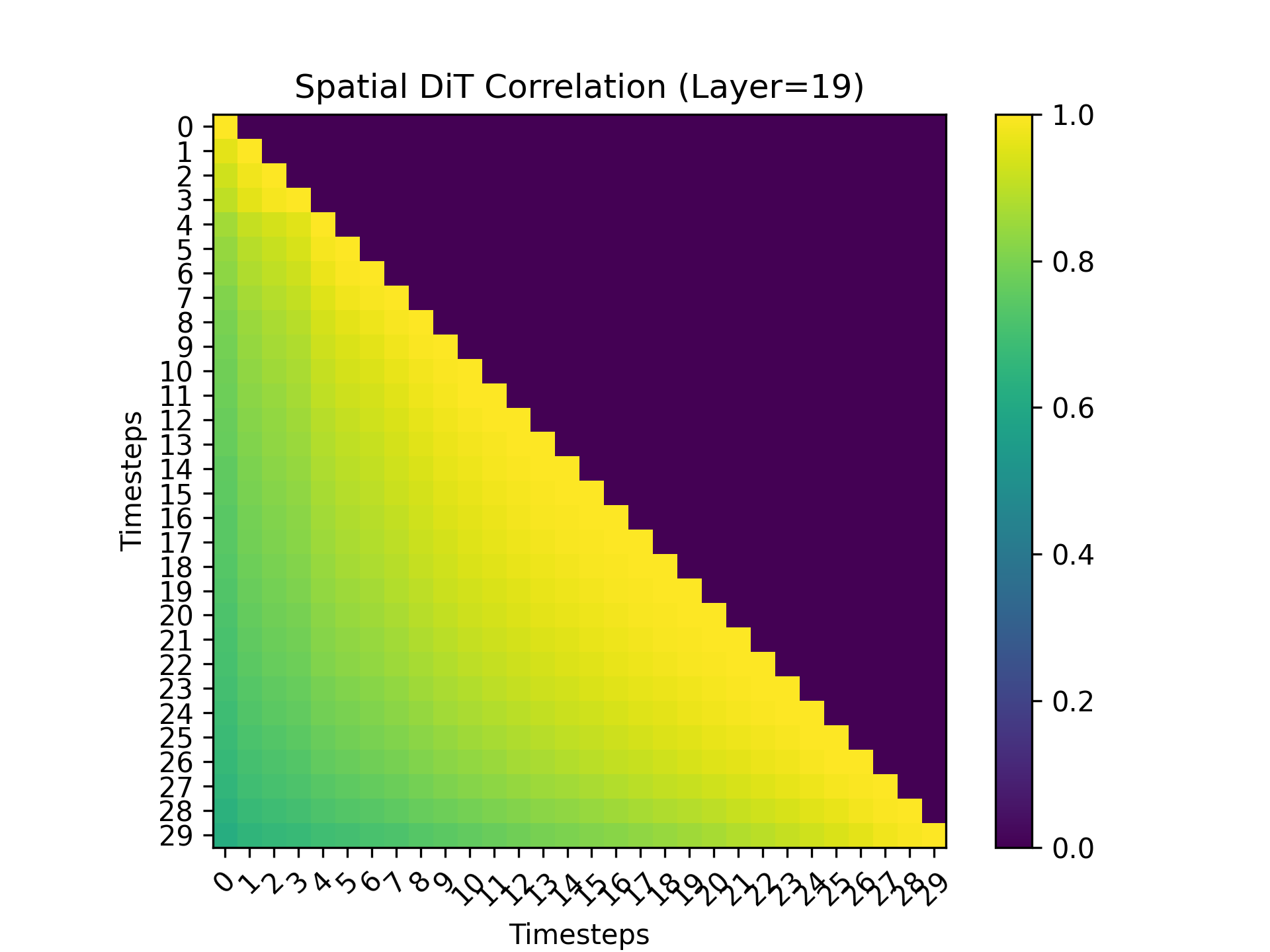}
  \end{minipage}
  }


\subfloat[$\mathrm{Layer\;24}$]{
  \begin{minipage}[t]{0.48\textwidth}
      \centering
      \includegraphics[width=\textwidth]{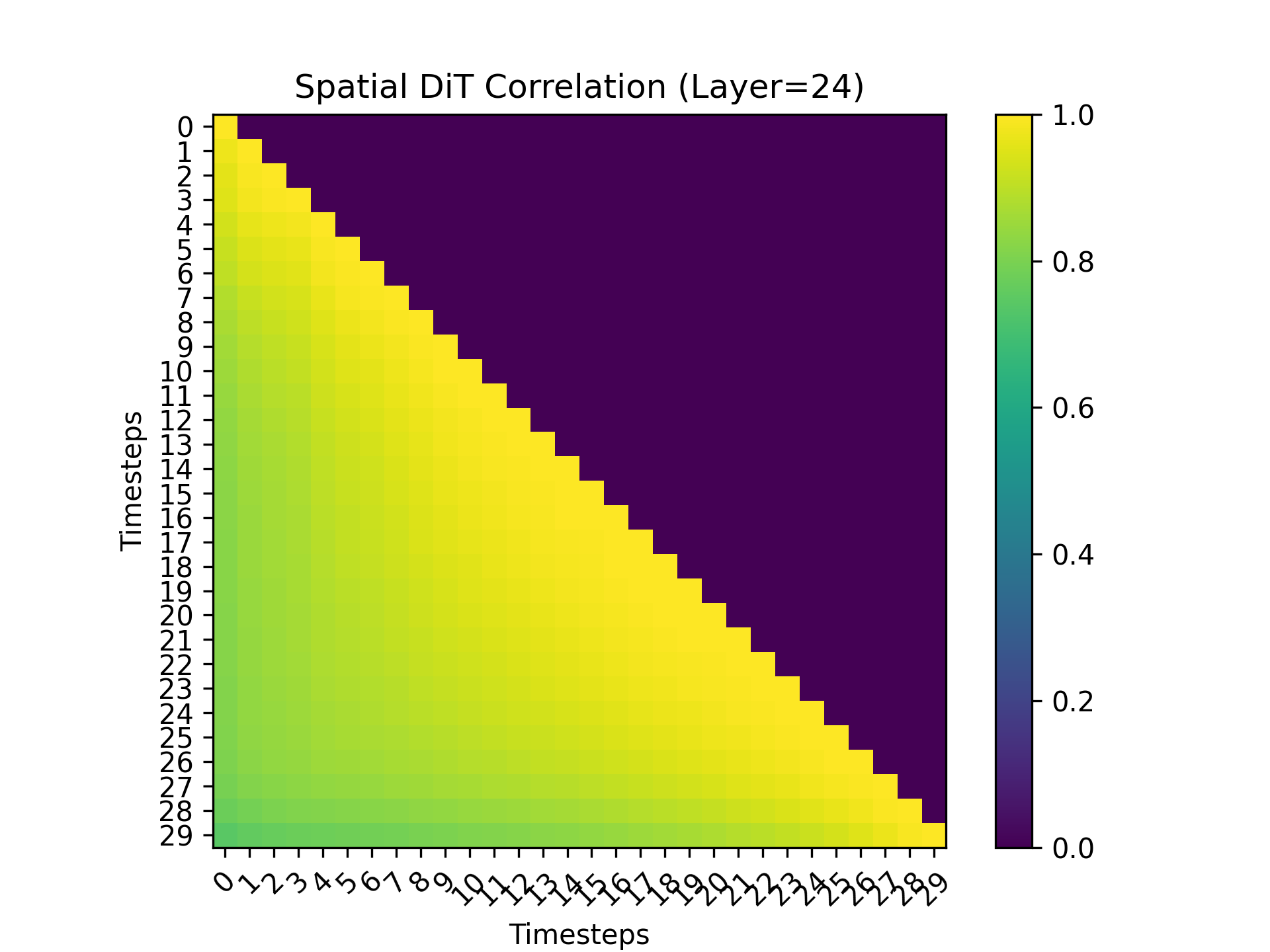}
  \end{minipage}
  }
  \hfill
  \subfloat[$\mathrm{Layer\;27}$]{
  \begin{minipage}[t]{0.48\textwidth}
  \centering
  \includegraphics[width=\textwidth]{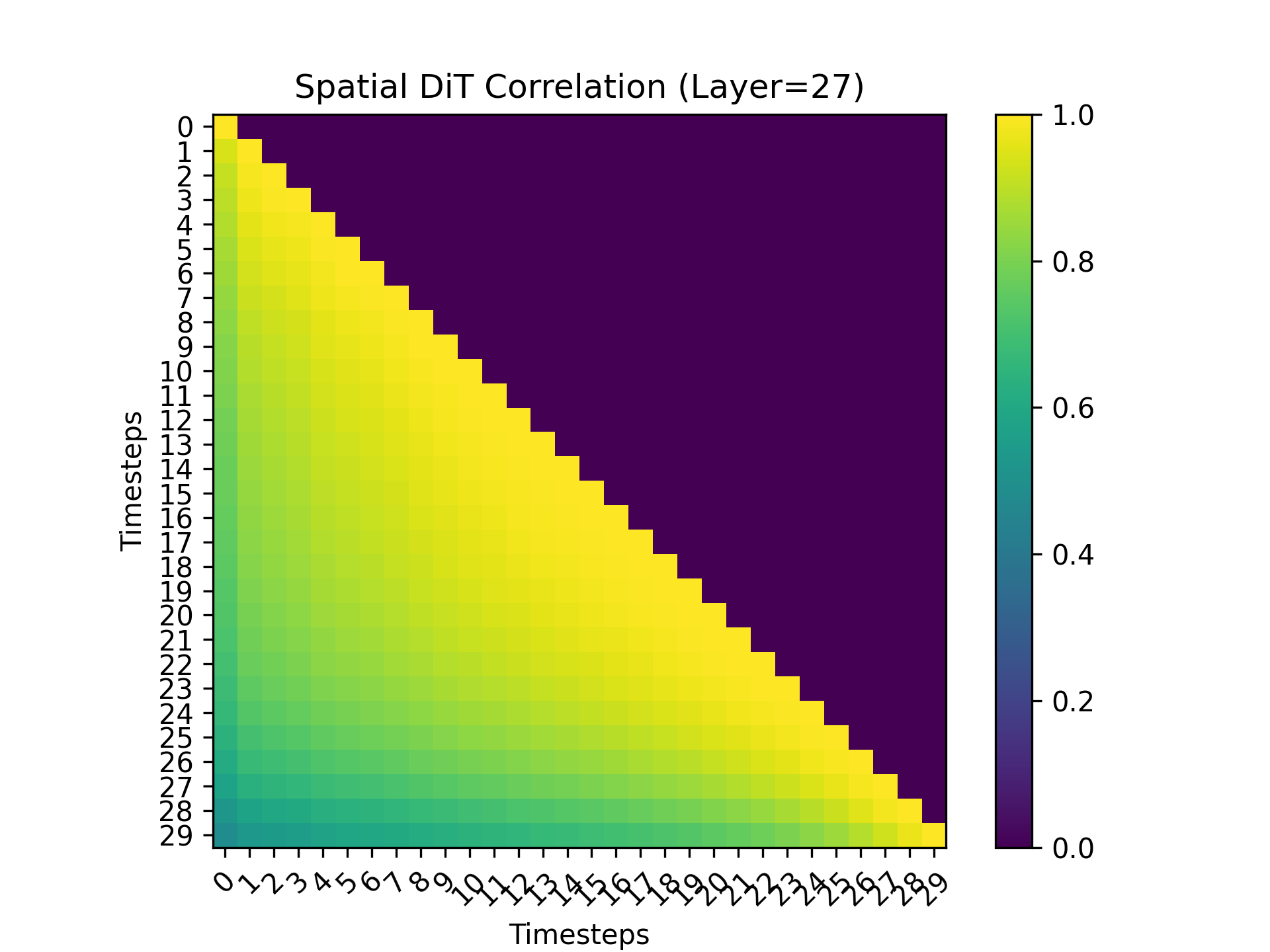}
  \end{minipage}
  }
  
  \caption{Cosine similarity of Spatial-DiT features across denoising steps for different layers of the OpenSora model.}
  \label{fig:across_denoising_steps}
\end{figure*}

Figure~\ref{fig:across_denoising_steps} shows the cosine similarity of Spatial-DiT features across denoising steps for different layers of the OpenSora model. Later layers exhibit greater feature variation than early and middle layers.

\section{Evaluation}
\subsection{Evaluation Metrics}
\label{subsec:eval_metrics}

We evaluate video generation quality using established metrics that assess both perceptual quality and similarity to baseline outputs. Specifically, we use VBench~\citep{vbench}, a comprehensive benchmark capturing multiple perceptual dimensions, and complement it with similarity metrics comparing generated videos to baseline outputs without reuse.

\paragraph{$\mathrm{VBench}$}~\citep{vbench} is a video generation benchmark that evaluates model quality across 16 well-defined dimensions. It uses 11 prompt categories, each designed to probe specific aspects of these dimensions, with weighted scores assigned to each. For evaluation, we select the first 50 prompts from each category, totaling 550 prompts.

\paragraph{$\mathrm{PSNR}$} measures pixel-level mean squared error between a baseline video (without reuse) and a reused-version. Higher PSNR indicates lower error and better quality.

\paragraph{$\mathrm{SSIM}$}~\citep{ssim} measures image similarity by comparing structural information, luminance, and contrast between two images.

\paragraph{$\mathrm{LPIPS}$}~\citep{lpips} measures perceptual similarity between images using deep neural network features, capturing differences more effectively than pixel-based metrics. It computes the distance between features extracted from pre-trained models trained on large-scale datasets.

\paragraph{$\mathrm{FVD}$}~\citep{fvd} based on the Fréchet Inception Distance (FID) for images, extends the concept to videos. It measures the distance between real and generated video distributions in the feature space of a pretrained network. By capturing both spatial quality and temporal consistency, FVD reflects how closely generated videos match real ones in appearance and motion.

We compute all the above similarity metrics per frame and report the average across all frames as the final video score.

\subsection{Baselines Generation Settings}
\label{subsec:baseline_config}

We evaluate four static cache reuse methods—$\mathrm{Static}$, $\Delta$-$\mathrm{DiT}$\citep{deltadit}, $\mathrm{T}$-$\mathrm{GATE}$\citep{tgate}, and $\mathrm{PAB}$\citep{pab}—using the configurations detailed below.
 
\paragraph{$\mathrm{Static}$} method caches coarse-grained features using a reuse window of size $N$ and updates the cache at a fixed compute interval $R$, as shown in Table~\ref{tab:baseline_static}.

\begin{table*}[h]
\centering
\caption{$\mathrm{Static}$ baseline Settings}
\input{./body/tables/baselines_static.tex}
\label{tab:baseline_static}
\end{table*}

\paragraph{$\Delta$-$\mathrm{DiT}$}

caches feature map deviations instead of full feature maps. It applies to back blocks during early outline generation and to front blocks during the detail refinement stage. Cache reuse is controlled by a gating hyperparameter $b$, which defines the boundary between front and back blocks, and a cache interval $k$ as shown in Table~\ref{tab:baseline_deltadit}.

\begin{table*}[h]
\centering
\caption{$\Delta$-$\mathrm{DiT}$ baseline Settings}
\input{./body/tables/baseline_deltadit.tex}
\label{tab:baseline_deltadit}
\end{table*}

\paragraph{$\mathrm{T}$-$\mathrm{GATE}$} divides the diffusion process into two phases: semantics planning and fidelity improvement, with the transition defined by gate step $m$. During the semantics-planning phase, cross-attention (CA) remains active, while self-attention (SA) is computed and reused every $k$ steps after an initial warm-up. After step $m$, CA is replaced by cached features, while SA continues as shown in Table~\ref{tab:baseline_tgate}.

\begin{table*}[h]
\centering
\caption{$\mathrm{T}$-$\mathrm{GATE}$ baseline Settings}
\input{./body/tables/baseline_tgate.tex}
\label{tab:baseline_tgate}
\end{table*}

\paragraph{$\mathrm{PAB}$} employs Pyramid Attention Broadcast, where the broadcast range forms a hierarchy: cross-attention $(\gamma)$ at the base, temporal attention $(\beta)$ in the middle, and spatial attention $(\alpha)$ at the top. Reuse occurs during designated broadcast timesteps. Each DiT block’s MLP follows a separate reuse schedule, as detailed in Table~\ref{tab:baseline_pab}, based on empirical evaluation.

\begin{table*}[h!]
\centering
\caption{$\mathrm{PAB}$ baseline Settings}
\input{./body/tables/baseline_pab.tex}
\label{tab:baseline_pab}
\end{table*}

\section{Extended Results}
\label{subsec:additional_results}

\subsection{Adaptive behavior for different prompts}

To quantify the adaptive behavior of \foresight{}, we plot absolute latency for all methods across prompts from the Open-Sora set. As shown in Figure~\ref{fig:latency_variation}, static reuse methods exhibit consistent latency due to fixed reuse schedules. In contrast, \foresight{} balances speed and quality based on scene complexity, enabling dynamic reuse for improved video quality and inference speedup.
Table~\ref{fig:latency_variation} shows the VBench wall clock time for 500 prompts for OpenSora model for a more realistic comparison.

\begin{figure}[h]
    \centering
    \begin{minipage}{0.48\textwidth}
        \centering
        \includegraphics[width=\linewidth]{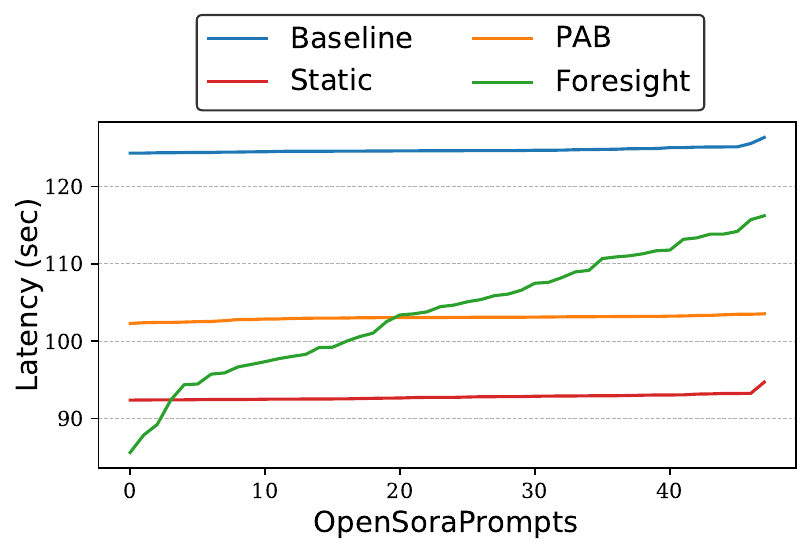}
        \label{fig:latency_variation}
    \end{minipage}\hfill
    \begin{minipage}{0.5\textwidth}
        \centering
        \resizebox{\linewidth}{!}{ 
        \begin{tabular}{l c}
        \toprule
        \multirow{2}{*}{\textbf{Method}} & \textbf{Cumulative Wall Clock Time} \\
         & \textbf{in seconds} (lower is better) \\
        \midrule
        $\mathrm{Baseline}$ & 7092.42 \\
        $\mathrm{PAB}$ & 5509.06 \\
        \foresight~($N_1R_2$) & 5455.77 \\
        \foresight~($N_2R_3$) & 4842.95 \\
        \bottomrule
        \end{tabular}}
        \label{tab:wall_clock_time}
    \end{minipage}
    \caption{\textbf{Left:} Latency variation across prompts from the Open-Sora set. Baseline and Static methods yield constant inference latency due to fixed reuse schedules. In contrast, \foresight{} adapts reuse based on prompt complexity, improving video quality with dynamic latency. Prompts are sorted by latency (ascending) using the Open-Sora model at 720p for 2-second generations. \textbf{Right:} Open-Sora cumulative wall clock time in seconds for VBench prompt set 500 prompts.}
    \label{fig:latency_variation}
\end{figure}

\subsection{Additional results}

To assess the efficacy of \foresight{}, we evaluate it on state-of-the-art models HunyuanVideo~\citep{hunyuanvideo} and Wan-2.1~\citep{wan2025} using the PenguinVideoBenchmark and compare it with TeaCache~\citep{teacache}. As shown in Table~\ref{tab:recent_comparison}, \foresight{} consistently outperforms prior work in quality while delivering competitive speedups. Unlike static methods, \foresight{} adapts to balance speed and quality. For fairness, we also report performance for the \foresight{} configuration that matches the quality of prior work.

\begin{table*}[h]
\centering
\input{./body/tables/recent_comparison.tex}
\caption{Qualitative comparison of \foresight{} with recent related work and state-of-the-art models on the PenguinVideoBenchmark. Results are shown for 10 randomly sampled prompts: 720p videos of 2 seconds with HunyuanVideo, and 480p videos of 5 seconds with Wan-2.1.}
\label{tab:recent_comparison}
\end{table*}

\subsection{Performance comparison with same generation quality}

\foresight{} uses fixed hyperparameters across all experiments and adapts consistently across models, prompts, and configurations. These parameters provide a practical way to balance quality and performance without additional tuning. By contrast, prior work~\citep{pab, teacache} requires model-specific adjustments: PAB fixes MLP reuse empirically, while TeaCache thresholds vary by model. For a fair comparison, Table~\ref{tab:perf_quality_comp} reports speedups for both methods using a \foresight{} configuration that matches PAB’s output quality across models.

\begin{table*}[h]
\centering
\caption{Performance improvement measured at matched video quality}
\input{./body/tables/perf_quality_comp.tex}
\label{tab:perf_quality_comp}
\end{table*}

\subsection{Additional results across other benchmarks and prompts set}

To complement Section~\ref{subsec:results}, which reports results using VBench~\citep{vbench} prompts and standard similarity metrics, we further evaluate \foresight{} on two additional prompt sets: UCF-101~\citep{ucf101} and EvalCrafter~\citep{evalcrafter}. We include $\mathrm{CLIPSIM}$ and $\mathrm{CLIP}$-$\mathrm{Temp}$ to assess text-video alignment and temporal consistency, along with DOVER’s~\citep{dover} $\mathrm{VQA}$ metrics for both aesthetic and technical quality, as shown in Table~\ref{tab:results_ucf}.

\begin{table*}[b!]
\centering
\input{./body/tables/results_ucf.tex}
\caption{Qualitative comparison of \foresight{} and PAB on UCF~\citep{ucf101} and EvalCrafter~\citep{evalcrafter} prompts set. Videos are generated at 240p with OpenSora, 512x512 with Latte, and 480x720 with CogVideoX, all with a fixed duration of 2 seconds. Metrics including EvalCrafter's~\cite{evalcrafter} $\mathrm{CLIP}$ score and DOVER's~\citep{dover} $\mathrm{VQA}$ in Aesthetics, Technical and Overall score.}
\label{tab:results_ucf}
\end{table*}

\subsection{Extension to other diffusion tasks}

To evaluate the efficacy of \foresight{} across other diffusion tasks, we apply \foresight{} to the FLUX~\citep{flux} text-to-image generation model, which does not include a temporal dimension. In this setting, \foresight{} reuses spatial and cross-attention layers and MLPs.

Table~\ref{tab:txt_to_img} reports the performance, showing that \foresight{} reduces wall-clock time by approximately $2\times$ while maintaining comparable image quality to the baseline.

\begin{table*}[h]
\centering
\caption{Open-Sora cumulative wall clock time in seconds}
\input{./body/tables/txt_to_img.tex}
\label{tab:txt_to_img}
\end{table*}

\subsection{Qualitative Comparison}

\begin{figure}[t]
  \centering
  \includegraphics[width=0.95\textwidth]{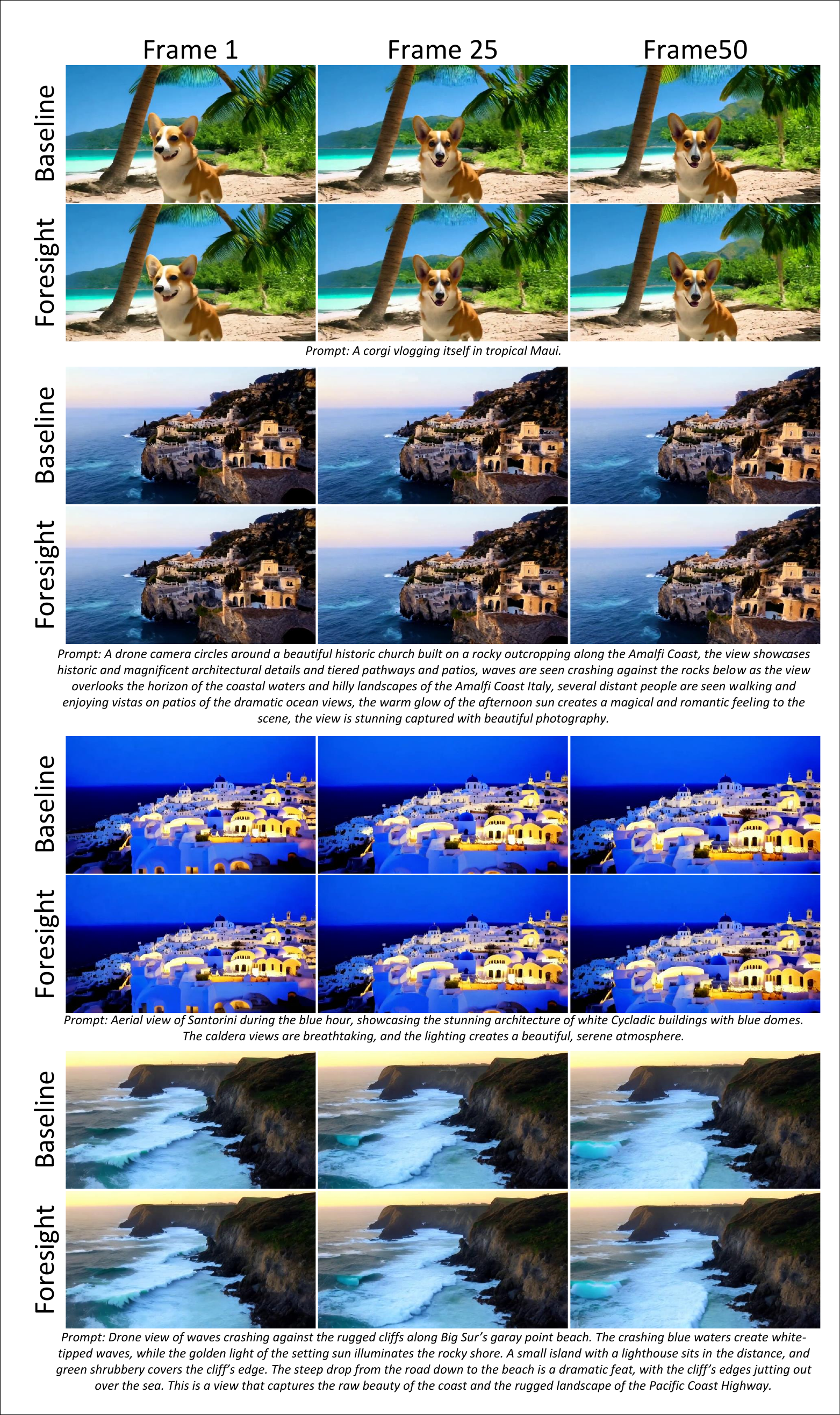}
  \caption{Qualitative Comparison for Open-Sora~\citep{opensora} with 720p, 2sec video generation across multiple frames.
}
\label{fig:opensora_visuals}
\end{figure}

\begin{figure}[t]
  \centering
  \includegraphics[width=0.95\textwidth]{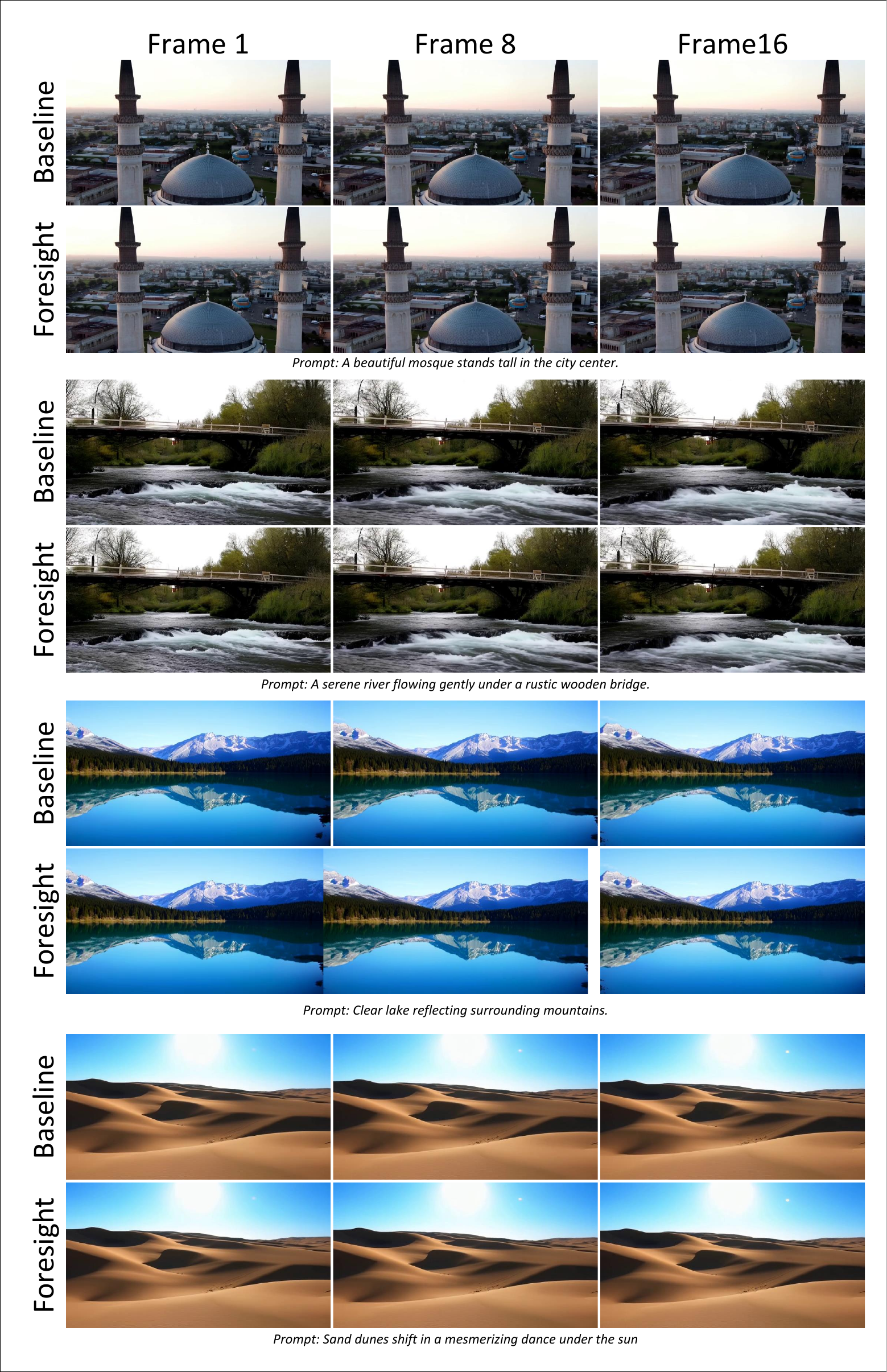}
  \caption{Qualitative Comparison for CogVideoX~\citep{cogvideox} with 720p, 2sec video generation across multiple frames.
}
\label{fig:cogvideox_visuals}
\end{figure}

Figures~\ref{fig:opensora_visuals} and~\ref{fig:cogvideox_visuals} present a qualitative comparison of videos generated by \foresight{} and the baseline across multiple frames. \foresight{} maintains consistent video quality while achieving notable performance improvements.

%% file: body/background.tex
\section{Background}
\label{sec:background}

The success of Sora~\cite{sora} has shown the strong potential of diffusion transformers for text-to-video generation, inspiring a wave of open-source models such as Open-Sora~\citep{opensora}, Open-Sora-Plan~\citep{opensora_plan}, Latte~\citep{latte}, CogVideoX~\citep{cogvideox}, Vchitect~\citep{vchitect}, and Mochi~\citep{mochi}.

\begin{figure*}[b]
  \centering
  \includegraphics[width=1\columnwidth]{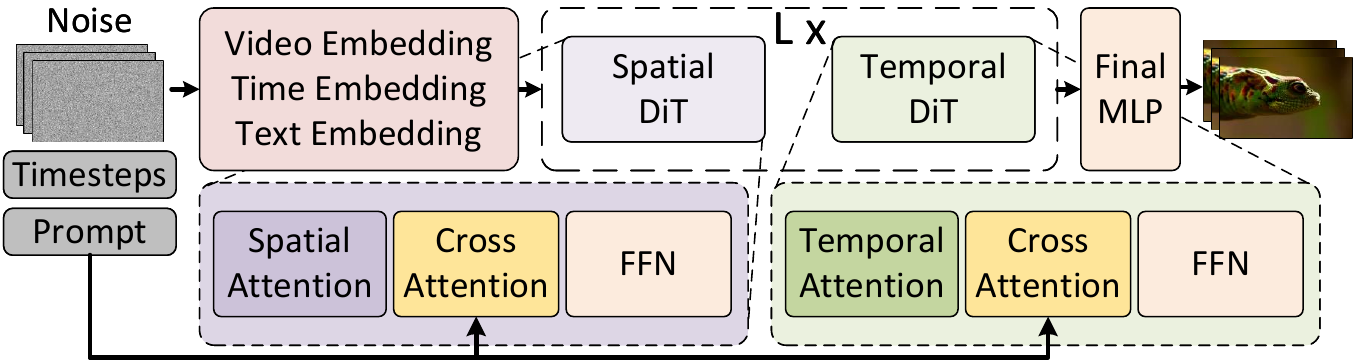}
  \caption{Overview of diffusion transformer (DiT) based text-to-video generation models. It comprises of mainly spatial and temporal diffusion transformer blocks with spatial and temporal attention. Cross attention takes into acccount prompt conditioning for generated video.
}
\label{fig:dit}
\end{figure*}

Figure~\ref{fig:dit} shows a high-level overview of DiT-based text-to-video models. These models take as input Gaussian noise-initialized latent video frames, timesteps for scheduling, and a text prompt. They use Spatial-Temporal DiT (ST-DiT) blocks: Spatial-DiT captures intra-frame spatial structure, while Temporal-DiT captures inter-frame temporal dynamics. Cross-attention injects text conditioning into each block, and feedforward networks (FFNs) apply learned nonlinearities. Timestep embeddings guide denoising across both spatial and temporal layers.

%% file: body/tables/baselines_static.tex
\resizebox{0.5\textwidth}{!}{
\begin{tabular}{l c c c}
\toprule
\multirow{2}{*}{$\mathrm{Static}$} & Denoising & Reuse  & Compute \\
 & Steps $(T)$ & Window $(N)$ & Interval $(R)$ \\
\midrule
Open-Sora & 30 & 1 & 2 \\
Latte & 50 & 1 & 2 \\
CogVideoX & 50 & 1 & 2 \\
\bottomrule
\end{tabular}
}

%% file: body/tables/baseline_deltadit.tex
\resizebox{0.55\textwidth}{!}{
\begin{tabular}{l c c c c}
\toprule
\multirow{2}{*}{$\Delta$-$\mathrm{DiT}$} & Denoising & Cache  & Gate & Block \\
 & Steps $(T)$ & Interval $(k)$ & Step $(b)$ & Range \\
\midrule
Open-Sora & 30 & 2 & 25 & [0, 5] \\
Latte & 50 & 2 & 48 & [0, 2] \\
CogVideoX & 50 & 2 & 48 & [0, 2] \\
\bottomrule
\end{tabular}
}

%% file: body/tables/baseline_tgate.tex
\resizebox{0.5\textwidth}{!}{
\begin{tabular}{l c c c}
\toprule
\multirow{2}{*}{$\mathrm{T}$-$\mathrm{GATE}$} & Denoising & Cache  & Gate \\
 & Steps $(T)$ & Interval $(k)$ & Step $(b)$ \\
\midrule
Open-Sora & 30 & 2 & 12 \\
Latte & 50 & 2 & 20 \\
CogVideoX & 50 & 2 & 20 \\
\bottomrule
\end{tabular}
}

%% file: body/tables/baseline_pab.tex
\resizebox{1\linewidth}{!}{
\begin{tabular}{l | c | c c c c | c c c}
\toprule
\multirow{2}{*}{$\mathrm{PAB}$} & Denoising & \multicolumn{4}{c|}{Broadcast Range and Timesteps} & \multicolumn{3}{c}{MLP Broadcast}  \\
 & Steps $(T)$ & Spatial $(\alpha)$ & Temporal $(\beta)$ & Cross $(\gamma)$ & Timesteps & Reuse &  Blocks & Timesteps \\
\midrule
Open-Sora & 30 & 2 & 4 & 6 & [930-450] & 2 & [0, 1, 2, 3, 4] & [864, 788, 676]] \\
Latte & 50 & 2 & 4 & 6 & [800-100] & 2 & [0, 1, 2, 3, 4] & [720, 640, 560, 480, 400]] \\
CogVideoX & 50 & 2 & - & - & [850, 100] & - & - & - \\
\bottomrule
\end{tabular}
}

%% file: body/tables/recent_comparison.tex
\resizebox{1\textwidth}{!}{
\begin{tabular}{l c c c c c c c c}
\toprule
\textbf{Model} & \textbf{Method} & $\mathrm{PSNR}\uparrow$ &  $\mathrm{SSIM}\uparrow$ & $\mathrm{LPIPS}\downarrow$ & $\mathrm{FVD}\downarrow$ & $\mathrm{Latency(s)}$ & $\mathrm{Speedup}\uparrow$
\\
\toprule
 \multirow{4}{*}{\textbf{HunyuanVideo}} & $\mathrm{Baseline}$ & - &	- &	- &	- &	415.94 ($\pm\;$1.04) & -
 \\
 & $\mathrm{TeaCache}(0.1)$ & 37.31 & 0.96 & 0.02 & 170.16 & 258.08 ($\pm\;$3.28) &	1.61$\times$
 \\
 & \foresight~($N_1R_2$) & 41.79 & 0.97 & 0.01 & 47.45 & 256.75 ($\pm\;$6.54) & 1.62$\times$
 \\
 & \foresight~($N_2R_3$) & 37.25 &	0.96 &	0.02 &	172.85 &	231.08 ($\pm\;$7.42) &	1.80$\times$
 \\
 \midrule
 \multirow{4}{*}{\textbf{Wan-2.1}} & $\mathrm{Baseline}$ & - & - & - &	- &	74.87 ($\pm\;$0.45) &	-
 \\
 & $\mathrm{TeaCache}(0.1)$ & 21.49 &	0.77 &	0.14 &	1222.56 &	38.9 ($\pm\;$0.25)	&	1.90$\times$
\\
 & \foresight~($N_1R_2$) & 	25.12 &	0.86 &	0.08 &	543.40 &	50.93 ($\pm\;$1.14) & 1.47$\times$
 \\
 & \foresight~($N_2R_3$) & 21.64 &	0.77 &	0.15 &	1234.48 &	33.57 ($\pm\;$1.32) &	2.23$\times$
\\
 \bottomrule
\end{tabular}}

%% file: body/tables/perf_quality_comp.tex
\resizebox{0.65\textwidth}{!}{
\begin{tabular}{l c c c}
\toprule
\textbf{Model} & $\mathrm{PSNR}$ & $\mathrm{PAB}$ Speedup & \foresight{} Speedup \\
\midrule
OpenSora & 25.67 & 1.26$\times$ & 1.68$\times$ \\
Latte & 21.22 & 1.29$\times$ & 1.58$\times$ \\
CogVideoX & 29.04 & 1.37$\times$ & 1.95$\times$ \\
\bottomrule
\end{tabular}
}

%% file: body/tables/results_ucf.tex
\resizebox{1\textwidth}{!}{
\begin{tabular}{l c c c c c c c c}
\toprule
\multirow{2}{*}{\textbf{Model}} & \multirow{2}{*}{\textbf{Method}} &
$\mathrm{CLIP}$ & $\mathrm{CLIP}$ &  $\mathrm{VQA}$ & $\mathrm{VQA}$ & $\mathrm{VQA}$ & \multirow{2}{*}{$\mathrm{Latency(s)}$} & \multirow{2}{*}{$\mathrm{Speedup}$}
\\
 & & $\mathrm{SIM}$ & $\mathrm{Temp}$ & $\mathrm{Aesthetic}$ & $\mathrm{Technical}$ & $\mathrm{Overall}$ & \\
\toprule
\multicolumn{9}{c}{\textbf{UCF-101~\citep{ucf101}} (101 Prompts)} \\
\toprule
 \multirow{4}{*}{\textbf{Open-Sora}} & $\mathrm{Baseline}$ & 20.51 &	99.64 &	14.56 &	25.02 &	19.75 &	12.71 ($\pm\;$0.04) &	- \\
 & $\mathrm{PAB}$ & 20.47 &	99.76 &	6.57 &	15.66 &	10.32 &	10.05 ($\pm\;$0.06) &	1.26$\times$ \\
 & \foresight~($N_1R_2$) & 20.73 &	99.88 &	15.17 &	26.68 &	21.11 &	9.72 ($\pm\;$0.84) &	1.30$\times$
 \\
 & \foresight~($N_2R_3$) & 20.70 &	99.86 &	14.01 &	26.11 &	20.32 &	7.32 ($\pm\;$0.80) &	1.73$\times$
 \\
 \midrule
 \multirow{4}{*}{\textbf{Latte}} & $\mathrm{Baseline}$ & 20.09 &	99.39 &	20.71 &	12.93 &	14.10 &	32.48 ($\pm\;$0.02)	& -
 \\
 & $\mathrm{PAB}$ & 20.09 &	99.39 &	20.71 &	12.69 &	13.79 &	25.11 ($\pm\;$0.01) &	1.29$\times$
\\
 & \foresight~($N_1R_2$) & 20.12 &	99.14 &	22.72 &	21.04 &	20.34 &	21.48 ($\pm\;$0.05) &	1.51$\times$
 \\
 & \foresight~($N_2R_3$) & 20.09 &	99.16 &	21.95 &	19.14 &	18.92 &	15.56 ($\pm\;$0.22) &	 2.08$\times$
\\
 \midrule
 \multirow{4}{*}{\textbf{CogVideoX}} & $\mathrm{Baseline}$ & 20.22 &	99.43 &	40.25 &	42.34 &	41.19 &	29.35 ($\pm\;$0.02)	& -
  \\
 & $\mathrm{PAB}$ & 20.20 &	99.43 &	39.44 &	41.86 &	40.52 &	22.65 ($\pm\;$0.05)	&	1.29$\times$
\\
 & \foresight~($N_1R_2$) & 20.21 &	99.42 &	40.64 &	40.65 &	40.17 &	19.57 ($\pm\;$0.96)	&	1.49$\times$
 \\
 & \foresight~($N_2R_3$) & 20.17 &	99.44 &	41.46 &	43.34 &	42.15 &	17.00 ($\pm\;$1.18) &	1.72$\times$
\\
 \bottomrule
 \multicolumn{9}{c}{\textbf{EvalCrafter~\citep{evalcrafter}} (150 prompts)} \\
 \toprule
 \multirow{4}{*}{\textbf{Open-Sora}} & $\mathrm{Baseline}$ & 20.07 &	99.55 &	17.48 &	29.94 &	23.49 &	12.59 ($\pm\;$0.09) & -
 \\
 & $\mathrm{PAB}$ & 19.91 &	99.76 &	9.16 &	20.82 &	14.15 &	10.07 ($\pm\;$0.03)	& 1.24$\times$
\\
 & \foresight~($N_1R_2$) & 20.05 &	99.54 &	16.85 &	29.56 &	23.10 &	9.36 ($\pm\;$0.77)	&	1.34$\times$
 \\
 & \foresight~($N_2R_3$) & 20.05 &	99.50 &	15.67 &	29.21 &	22.24 &	7.06 ($\pm\;$0.93)	& 1.78$\times$
 \\
 \midrule
 \multirow{4}{*}{\textbf{Latte}} & $\mathrm{Baseline}$ & 20.69 &	99.50 &	53.05 &	45.59 &	48.37 &	32.48 ($\pm\;$0.02)	& -
 \\
 & $\mathrm{PAB}$ & 19.98 &	99.65 &	53.35 &	35.39 &	40.84 &	25.11 ($\pm\;$0.01) &	1.29$\times$
\\
 & \foresight~($N_1R_2$) & 20.60 &	99.51 &	55.85 &	45.80 &	49.27 &	28.37 ($\pm\;$1.14)	&	1.14$\times$ 
 \\
 & \foresight~($N_2R_3$) & 20.54 &	99.50 &	54.00 &	44.03 &	47.29 &	25.59 ($\pm\;$1.36)	&	1.26$\times$
\\
 \midrule
 \multirow{4}{*}{\textbf{CogVideoX}} & $\mathrm{Baseline}$ & 20.66 &	99.51 &	54.72 &	57.67 &	56.53 &	30.43 ($\pm\;$0.12)	&	-
  \\
 & $\mathrm{PAB}$ & 20.66 &	99.51 &	52.49 &	56.10 &	54.85 &	22.95 ($\pm\;$0.07) &	1.32$\times$
\\
 & \foresight~($N_1R_2$) & 20.66 &	99.55 &	53.34 &	56.73 &	55.64 &	25.80 ($\pm\;$0.84) &	1.17$\times$
 \\
 & \foresight~($N_2R_3$) & 20.64 &	99.55 &	52.54 &	56.34 &	54.98 &	24.56 ($\pm\;$1.08)	&	1.23$\times$
\\
 \bottomrule
\end{tabular}}

%% file: body/tables/txt_to_img.tex
\resizebox{0.45\textwidth}{!}{
\begin{tabular}{l c}
\toprule
\textbf{Method} & \textbf{Latency (seconds)} \\
\midrule
$\mathrm{Baseline}$ & 14.02 \\
\foresight~($N_1R_2$) & 7.1 \\
\bottomrule
\end{tabular}
}